\documentclass{ieeeaccess}
\usepackage{cite}
\usepackage{amsmath,amssymb,amsfonts}
\usepackage{algorithmic}
\usepackage{graphicx}
\usepackage{textcomp}

\usepackage{pifont}
\usepackage{comment}
\usepackage{multirow}
\usepackage{multicol}
\usepackage{colortbl}
\usepackage{booktabs}
\usepackage{hyperref}
\usepackage{url}
\usepackage[capitalize]{cleveref}
\usepackage{bookmark}
\usepackage{algorithm}
\usepackage{bm}

\usepackage{bm}
\makeatletter
\renewcommand{\iheaderfont}{\sffamily\fontencoding{T1}\fontseries{m}\fontshape{it}\fontsize{7}{7}\selectfont}
\AtBeginDocument{\DeclareMathVersion{bold}
\SetSymbolFont{operators}{bold}{T1}{times}{b}{n}
\SetSymbolFont{NewLetters}{bold}{T1}{times}{b}{it}
\SetMathAlphabet{\mathrm}{bold}{T1}{times}{b}{n}
\SetMathAlphabet{\mathit}{bold}{T1}{times}{b}{it}
\SetMathAlphabet{\mathbf}{bold}{T1}{times}{b}{n}
\SetMathAlphabet{\mathtt}{bold}{OT1}{pcr}{b}{n}
\SetSymbolFont{symbols}{bold}{OMS}{cmsy}{b}{n}
\renewcommand\boldmath{\@nomath\boldmath\mathversion{bold}}}
\makeatother

\def\BibTeX{{\rm B\kern-.05em{\sc i\kern-.025em b}\kern-.08em
    T\kern-.1667em\lower.7ex\hbox{E}\kern-.125emX}}
\definecolor{gray11}{gray}{0.9}

\usepackage{xcolor}

\begin{document}
\history{Date of publication xxxx 00, 0000, date of current version xxxx 00, 0000.}
\doi{}

\title{EventEgoHands++: Event-based Egocentric
3D Hand Mesh Reconstruction with Real Dataset}

\author{
\uppercase{Ryosei Hara}\authorrefmark{1}, \IEEEmembership{Graduate Student Member, IEEE}, \\
\uppercase{Wataru Ikeda}\authorrefmark{1}, \IEEEmembership{Graduate Student Member, IEEE}, \\
\uppercase{Masashi Hatano}\authorrefmark{1}, \IEEEmembership{Graduate Student Member, IEEE}, \\
\uppercase{Mariko Isogawa}\authorrefmark{1, 2}, \IEEEmembership{Member, IEEE}
}

\address[1]{Graduate School of Science and Technology, Keio University, Yokohama, Kanagawa 223-8522, Japan}
\address[2]{JST Presto}
\tfootnote{
This research is supported by JST Presto JPMJPR22C1, JSPS KAKENHI 24K22296, and 25H01159, JST FOREST Program JPMJFR242I.
W. Ikeda is supported by JST BOOST, Japan Grant Number JPMJBS2409.
M. Hatano is supported by JST BOOST, Japan Grant Number JPMJBS2409, and Amano Institute of Technology.
}

\markboth
{R. Hara \headeretal: EventEgoHands++: Event-based Egocentric
3D Hand Mesh Reconstruction with Real Dataset}
{R. Hara \headeretal: EventEgoHands++: Event-based Egocentric
3D Hand Mesh Reconstruction with Real Dataset}

\corresp{Corresponding author: Ryosei Hara (e-mail: ryosei\_hara@keio.jp).}

\begin{abstract}
3D hand mesh reconstruction is a challenging yet essential task for downstream applications, including human-robot interaction and AR/VR. 
Although conventional cameras (\eg, RGB or depth cameras) have been widely adopted for this task, methods that rely on them struggle in low-light environments and under severe motion blur.
To address these limitations, event-based cameras have recently attracted attention for their high dynamic range and high temporal resolution.
However, applying event cameras to egocentric hand reconstruction remains challenging because camera wearer's motion produces dense background events that obscure hand-specific signals.
Although the first egocentric event-based approach mitigates this issue using hand segmentation, its binary hand mask does not distinguish between left and right hands.
As a result, the model lacks instance-level hand information and predicts both hands even when only one or neither hand is present.
This limitation leads to incorrect inter-hand relationships and degraded reconstruction accuracy.
In this paper, we propose EventEgoHands++, a framework for event-based 3D hand mesh reconstruction from an egocentric viewpoint that overcomes these limitations.
The proposed method incorporates a Hand Detector that estimates instance-level bounding boxes and masks for both the left and right hands.
Moreover, we introduce Adaptive Attention, which dynamically gates the attention based on these detection results to accurately learn the spatial relationship and mutual interactions between the hands.
To train and evaluate our framework, we extend the synthetic
N-HOT3D dataset with bounding-box annotations and refined masks, and
newly construct EEH-R, the largest real-world event-based egocentric hand dataset to date, comprising approximately 1M annotated frames captured in environments including low-light conditions.
Extensive experiments on both synthetic and real datasets demonstrate that our method consistently outperforms the baselines.
Our code and datasets are available at \url{https://ryhara.github.io/EventEgoHandsV2/}.
\end{abstract}

% Enter key words or phrases in alphabetical order, separated by commas.
 % Autocorrelation, beamforming, communications technology, dictionary learning, feedback, fMRI, mmWave, multipath, system design, multipath, slight fault, underlubrication fault.

\titlepgskip=-21pt

\maketitle

%================================================================
\section{Introduction}
\label{sec:introduction}

\PARstart{H}{ands} play a fundamental role in human interaction with the physical world.
For example, daily activities such as grasping objects, manipulating tools, and performing gestures rely heavily on hand motions. 
Therefore, reconstructing a 3D hand mesh from egocentric vision is important for applications such as human–computer interaction, AR/VR, and robotics.
In particular, capturing fine-grained hand motions and shapes from the camera wearer’s viewpoint is essential for enabling immersive experiences and safe interactions.

However, conventional methods based on RGB(D) cameras~\cite{ge2016robust, xu2013efficient, ge2018real, boukhayma20193d, pavlakos2024hamer, potamias2025wilor} may encounter challenging situations when applied to egocentric vision.
For example, in addition to rapid hand movements, the head motion of the camera wearer can cause motion blur.
Moreover, in environments with varying illumination conditions, particularly in low-light settings, recognizing the hand can become challenging.

Recently, the use of event-based cameras, hereafter referred to as event cameras, has attracted increasing attention~\cite{gallego2022eventsurvey}.
Event cameras provide high temporal resolution and a high dynamic range, enabling the capture of fast motions even in low-light environments where RGB cameras struggle.
They also offer low power consumption and memory-efficient sensing, making them well suited for deployment on edge devices and wearable platforms.

\begin{figure}[t]
  \centering
  \includegraphics[width=\linewidth]{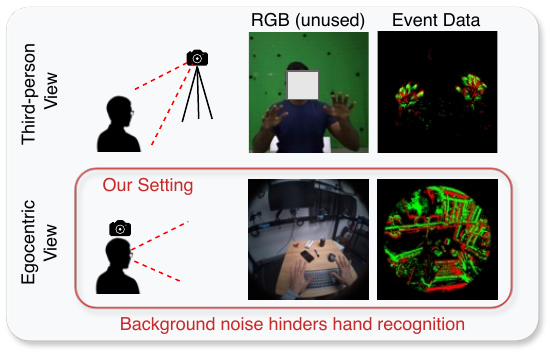}
  \vspace{-2mm}
  \caption{Challenges of egocentric event-based cameras. 
  While fixed third-person camera setups are limited to specific environments, egocentric viewpoints provide greater mobility and flexibility for capturing hand interactions. 
  However, in egocentric event camera scenarios, the wearer’s motion generates numerous background events, which can obscure hand-related signals and make accurate hand recognition difficult.}
  \label{fig:teaser}
\end{figure}

However, most existing event-based 3D hand mesh reconstruction methods~\cite{rudnev2021eventhands, jiang2024evhandpose, millerdurai2024ev2hands, liu2025leveraging} focus on fixed third-person camera setups.
In contrast, an egocentric perspective provides greater flexibility and mobility for capturing natural hand interactions, as illustrated in \cref{fig:teaser}.
One of the main challenges in using an event camera in a first-person setting is that camera motion generates a large number of events across the entire background, making it difficult to extract events corresponding to hand motion.
In our earlier conference paper, we proposed
EventEgoHands~\cite{Hara2025EventEgoHands}, the first framework for
egocentric event-based hand reconstruction, which suppresses such
background events with a hand segmentation module and established the
feasibility of this task.

Nevertheless, EventEgoHands should be regarded as a first step,
as it leaves three fundamental problems unresolved.

\noindent\textbf{Lack of instance-level hand identity.}
The binary mask in~\cite{Hara2025EventEgoHands} does not encode left/right hand identity, so the model always reconstructs both hands regardless of
their actual visibility, which frequently breaks in egocentric video
where hands often leave the field of view.
This yields invalid outputs when only one or neither hand is present
and degrades the estimated inter-hand relative positions.

\noindent\textbf{Visibility agnostic interaction modeling.}
Its cross-attention is applied unconditionally between the two hand
branches, forcing feature exchange with absent or misidentified hands.
Such a fixed attention pathway cannot adapt to the constantly changing
hand visibility inherent to egocentric interaction.

\noindent\textbf{Unverified real-world applicability.}
Its validation was confined to synthetic data, although real event
streams differ substantially from simulated ones in event density,
temporal distribution, and sensor noise.
Since no real-world egocentric event dataset existed, the effectiveness
of this task on real sensor data remained unverified.
In particular, performance under low-light conditions, the very
scenario that motivates event cameras, had never been evaluated on
real data.

To address these problems, we propose \textbf{EventEgoHands++}, a robust framework for event-based egocentric 3D hand mesh reconstruction.
Our proposed method consists of two main stages: hand extraction and hand reconstruction.
First, to resolve the inability to distinguish between left and right hands, we incorporate a Hand Detector in the hand extraction stage.
This detector simultaneously estimates instance-level bounding boxes and masks, ensuring that each hand is accurately localized and identified only when present.
Second, to address the performance degradation caused by fixed attention mechanisms, we introduce Adaptive Attention in the hand reconstruction stage. 
Unlike existing methods that apply cross-attention regardless of visibility, our Adaptive Attention dynamically selects the attention operations
according to the detection results.
This allows the model to effectively learn spatial correlations and inter-hand interactions by adapting to the visibility of each hand.

In addition, to facilitate research in this field, we provide two datasets.
Specifically, we extend \textbf{N-HOT3D}, the synthetic dataset
introduced in ~\cite{Hara2025EventEgoHands},
with refined segmentation masks and newly added bounding-box
annotations, enlarging it to approximately 480K annotated frames.
More importantly, we introduce \textbf{EEH-R}, a newly collected real-world dataset captured with an actual event camera in challenging egocentric scenarios, including low-light conditions.
While event-based hand reconstruction on real event data has been explored only in third-person settings, EEH-R enables, for the first time, evaluation on real event data in egocentric settings.
To the best of our knowledge, EEH-R is the largest real-world event-based egocentric hand dataset, containing approximately 1M annotated frames.
The details of the two datasets, N-HOT3D and EEH-R, are summarized in Table \ref{tab:event_hand_methods_rows}.

Using both datasets, we validate the effectiveness of EventEgoHands++. 
On the synthetic N-HOT3D dataset, our method reduces mean MPJPE by 21.82 mm (33.7\%) and MPVPE by 20.57 mm (34.0\%) compared with the best existing method. 
On the EEH-R dataset, it further reduces MPJPE by 7.90 mm (18.8\%) and MPVPE by 7.13 mm (18.1\%).

%================================================================
\begin{table*}[t]
\caption{Comparison of event-based hand datasets. The asterisk indicates that the corresponding annotation is available for a subset of the frames.}
\label{tab:event_hand_methods_rows}
\centering
\small
\setlength{\tabcolsep}{6pt}
\begin{tabular}{l|l|c|c|c|c|c|c|c}
\hline
Type & Dataset & View & Hands & Subs. & Dur. & Seq. & \#GT & Annotation \\
\hline
\hline

\multirow{3}{*}{Synthetic}
& EventHands (Synthetic)~\cite{rudnev2021eventhands} & Third & Single & - & 100 h & - & 360M & MANO  \\
& Ev2Hands-S~\cite{millerdurai2024ev2hands} & Third & Both & - & - & - & 28K & MANO, Event seg. \\
& \cellcolor{gray11}\textbf{N-HOT3D (Ours)} 
& \cellcolor{gray11}\textbf{Ego} 
& \cellcolor{gray11}Both 
& \cellcolor{gray11} \textbf{9} 
& \cellcolor{gray11} 4.4 h 
& \cellcolor{gray11} \textbf{136} 
& \cellcolor{gray11} 480K 
& \cellcolor{gray11} MANO, Hand mask/bbox \\

\hline

\multirow{4}{*}{Real}
& EventHands (Real)~\cite{rudnev2021eventhands}& Third & Single & - & 12.6 s & 4 & 357 & 2D keypoints \\
& Ev2Hands-R~\cite{millerdurai2024ev2hands} & Third & Both & 5 & 20.1 m & 8 & 70K & 3D keypoints \\
& EvRealHands~\cite{jiang2024evhandpose} & Third & Single & 10 & 79 m & 102 & 425K & MANO \\
& \cellcolor{gray11}\textbf{EEH-R (Ours)} 
& \cellcolor{gray11}\textbf{Ego} 
& \cellcolor{gray11}Both 
& \cellcolor{gray11}8 
& \cellcolor{gray11} \textbf{2.36 h} 
& \cellcolor{gray11}85 
& \cellcolor{gray11}\textbf{1M} 
& \cellcolor{gray11}MANO, (Hand mask/bbox)* \\

\hline
\end{tabular}
\vspace{-2mm}
\end{table*}

%================================================================

Our technical contributions are summarized as follows:
\begin{itemize}

\item We propose \textbf{EventEgoHands++}, a robust event-based egocentric 3D hand mesh reconstruction framework that explicitly handles hand visibility and left/right hand identity under severe egocentric camera motion.

\item We introduce two key methodological components: a \textbf{Hand Detector} that jointly estimates instance-level bounding boxes and masks for the left and right hands, and \textbf{Adaptive Attention}, which dynamically activates inter-hand attention based on detection results to model spatial relationships only when the corresponding hands are visible.

\item We newly construct \textbf{EEH-R}, the first and largest
real-world event-based egocentric hand dataset, containing approximately
1M annotated frames captured under both well-lit and low-light
conditions. We further extend the synthetic \textbf{N-HOT3D}
dataset with refined masks and newly added
bounding-box annotations, enlarging it to approximately 480K annotated
frames. Experiments on these datasets demonstrate the effectiveness of EventEgoHands++ against existing baselines.

\end{itemize}

The paper is structured as follows.
Section~\ref{sec:related_work} reviews related work on egocentric hand analysis, 3D hand pose estimation, and event-based 3D hand mesh reconstruction.
Section~\ref{sec:proposed_method} describes the proposed EventEgoHands++ framework, including the hand extraction and hand reconstruction stages.
Section~\ref{sec:dataset_collection} introduces the synthetic N-HOT3D dataset and the real-world EEH-R dataset.
Section~\ref{sec:experiment} presents the experimental setup, including implementation details, baseline methods, and evaluation metrics.
Section~\ref{sec:results} reports the experimental results, including evaluations on both datasets, hand segmentation performance, ablation studies, and failure case analysis.
Finally, Section~\ref{sec:conclusion} concludes the paper.

%================================================================

\section{Related Work}
\label{sec:related_work}

\subsection{Hands in Egocentric Videos}
\label{ssec:hands_ego}
Hands are the primary interface where humans interact with the world, making their analysis central to egocentric (first-person) video understanding~\cite{bandini2023analysis, plizzari2024outlook}.
Unlike exocentric (third-person) perspectives, egocentric video presents unique challenges, including frequent motion blur and severe occlusions during manipulation.

% hand-object interaction
Early research in egocentric hand analysis focused on localization and segmentation~\cite{fathi2011learning, lee2014this, Li_2013_ICCV, Li_2013_CVPR}.
Bambach~\etal~\cite{Bambach_2015_ICCV} introduced the EgoHands dataset, establishing a hands detection benchmark.
The research focus has recently extended from hand-centric tasks toward full hand-object interaction understanding with the advent of the large-scale interaction understanding using massive benchmarks such as EPIC-KITCHENS~\cite{Damen2018EPICKITCHENS, Damen2021PAMI}, Ego4D~\cite{Grauman_2022_CVPR}, and HOI4D~\cite{liu2022hoi4d}.
Recent efforts prioritize hand-object detector~\cite{cheng2023towards, Shan20}, joint hand-object reconstruction~\cite{yuan2025self, zhu2025rohit}, and hand-object segmentation~\cite{VISOR, zhang2022fine}.

% hands for downstream application
In addition to understanding hand-object interaction, egocentric hand cues serve as a vital proxy for downstream applications such as action recognition~\cite{Hatano2024MMCDFSL, zhanghelpinghand, liu2022joint} and anticipation~\cite{liu2019forecasting}.
Beyond the categorization of current and future actions, researchers have also focused on the temporal evolution of hand motion through hand forecasting~\cite{Hatano2024EMAG, Hatano2025EgoH4, ma2025mmtwin, ma2025diff-ip2d, ma2025madiff, Bao_2023_ICCV}.
These tasks are essential for proactive human-robot collaboration and augmented reality interfaces.

% hand pose estimation datasets (FPHA, H2O, HoloAssist, Assembly101, Ego-Exo4D, HOT3D, ARCTIC, DexYCB) & methods (EgoHandICL, AssemblyHands, S2D, HTT)
As a critical subfield, egocentric hand pose estimation has evolved alongside a progression of increasingly complex datasets.
While early benchmarks like FPHA~\cite{FirstPersonAction_CVPR2018} laid the foundation, recent datasets like H2O~\cite{Kwon_2021_ICCV}, HoloAssist~\cite{HoloAssist2023}, ARCTIC~\cite{fan2023arctic}, and Assembly101~\cite{sener2022assembly101} have introduced challenges involving bimanual interaction and articulated objects.
The state-of-the-art has been further pushed by large-scale datasets such as Ego-Exo4D~\cite{Grauman_2024_CVPR} and HOT3D~\cite{banerjee2025hot3d}. 
Methodologically, recent literature has shifted toward leveraging Transformers for spatial-temporal dependencies, as seen in HTT~\cite{wen2023hierarchical}, utilizing multi-view consistency to resolve egocentric occlusions, exemplified by AssemblyHands~\cite{ohkawa2023AssemblyHands} and S2DHand~\cite{liu2024single}, and addressing the task in the in-context learning~\cite{xie2026egohandicl}.
A more general literature review of 3D hand pose estimation follows next.

\subsection{3D Hand Pose Estimation}
\label{ssec:3d_hand}
% literature
Initial efforts in 3D hand pose estimation primarily relied on depth sensors~\cite{ge2016robust, ge2018real, xu2013efficient} to capture the complex geometry of the hand.
With the shift toward monocular RGB input, the field has largely converged on the use of parametric models to provide anatomical priors.
The MANO model~\cite{romero2017MANO} has become the de facto standard in this field, offering a low-dimensional yet anatomically plausible representation of hand shape and pose. 
Boukhayma~\etal~\cite{boukhayma20193d} pioneered the first fully learnable pipeline to directly regress MANO parameters from a single image, followed by the use of intermediate representations like 2D heatmaps~\cite{zhang2019end}, iterative 2D alignment loops~\cite{baek2019pushing}, and occlusion-robust methods~\cite{Park_2022_CVPR}.

% non-parametric
On the other hand, non-parametric methods~\cite{chen2022mobrecon, dkulon2019rec, kulon2020weakly} have explored direct vertex regression using Graph Convolutional Networks (GCNs)~\cite{kipf2017gcn}, which have been widely used to exploit the inherent graph structure of the hand mesh. THOR-Net~\cite{aboukhadra2023thor} further extended these to hand-object interaction, in which two hands and object poses are estimated.
More recently, Transformer-based architectures, such as METRO~\cite{lin2021end} and Mesh Graphormer~\cite{lin2021mesh}, have set new state-of-the-art benchmarks by modeling global interactions between vertices and joints without relying on a fixed kinematic tree.

% parametric (HaMeR, WiLoR)
However, the parametric approach remains highly favored for its robustness to occlusions and its ability to maintain valid hand topology. 
Recently, Pavlakos~\etal~\cite{pavlakos2024hamer} demonstrated that the performance of parametric reconstruction can be significantly enhanced by scaling up model capacity based on Vision Transformer (ViT)~\cite{dosovitskiy2021vit} backbones, following the success in body pose estimation~\cite{cai2024smplerx, goel2023humans}.
Recent image-based 3D hand pose estimation methods~\cite{pavlakos2024hamer, potamias2025wilor} achieve state-of-the-art accuracy across diverse datasets, and their encoders have also been shown to be useful for downstream tasks such as visibility estimation~\cite{hara2026handvisibilitydetector}.
Several works~\cite{ye2026prediciting4d, Zhang_2025_CVPR, yu2025dynhamr} have extended their efforts to 4D hand mesh reconstruction that incorporates temporal information.

While monocular 3D hand mesh reconstruction has advanced significantly, conventional camera-based approaches often fail in low-light conditions and under motion blur caused by rapid hand or camera movement. To this end, this study explores the use of event cameras, which are robust in such extreme scenarios due to their high dynamic range and high temporal resolution.

\begin{figure*}[t]
  \centering
  \includegraphics[width=\textwidth]{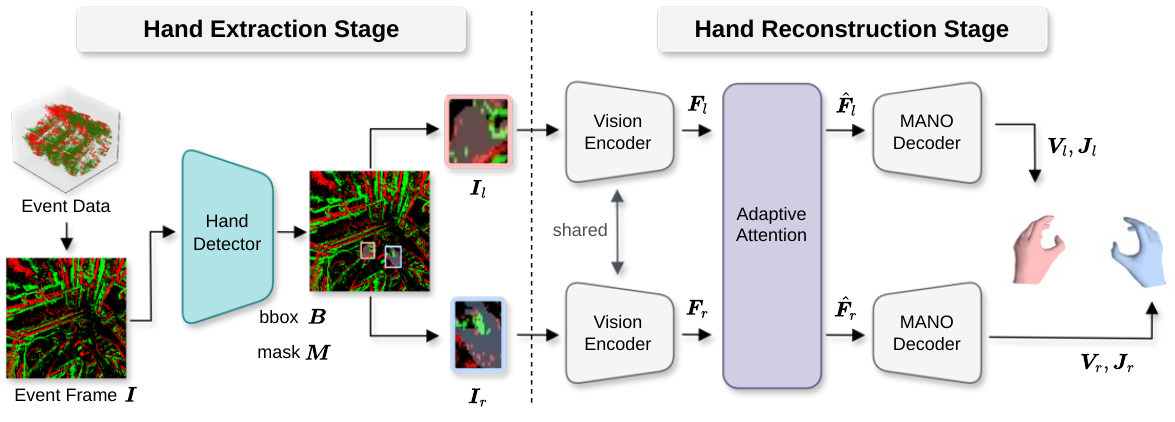}
  \caption{Architecture of EventEgoHands++. The pipeline consists of two primary stages: (1) Hand Extraction Stage, where a hand detector identifies and extracts hand regions from event data, and (2) Hand Reconstruction Stage, which utilizes a vision encoder for feature extraction and estimates 3D hand poses through Adaptive Attention based on the detection results.}
  \label{fig:model}
\end{figure*}

\subsection{Event-based 3D Hand Mesh Reconstruction}
\label{ssec:event_based_3d_hand}

% general story about event cameras
Event cameras output asynchronous streams of events by recording changes in pixel intensity.
In recent years, event cameras have gained increasing popularity~\cite{gallego2022eventsurvey, Xiao_2025_CVPR, Yura_2025_CVPR, Maeda_2025_CVPR, geckeler2026event} owing to their high dynamic range and high temporal resolution, which help address challenges such as low-light conditions and motion blur that conventional sensors, such as RGB or depth cameras, often fail to handle effectively.

% 3D reconstruction
In the context of 3D vision, event cameras are particularly advantageous as they provide nearly continuous trajectories of moving edges, allowing for the recovery of precise geometric structures and motion cues that are typically lost during fast movements in frame-based captures.
Recently, event cameras have been utilized for several 3D tasks~\cite{kar2025event}, such as 3D geometric reconstruction~\cite{rudnev2023eventnerf, Nakabayashi_2025_ev4dgs, deguchi2024e2gs, yu2025evagaussians}, non-rigid reconstruction~\cite{Xue_2022_BMVC}, depth estimation~\cite{ghosh2025event, horikawa2025dense}, human pose estimation~\cite{Ikeda2025DEventEgo, millerdurai2024eventego3d, millerdurai2024eventego3d2025eventego3dplusplus, hori2025eventpointmesh}, and 3D hand tracking~\cite{xu2025evhand}.

3D hand mesh reconstruction is a central task for various applications, such as robotics or AR/VR. Some studies~\cite{park20243d, jiang2024complementing} have utilized both RGB and event data to complement modality-specific information. Although these methods are robust, the hardware setup for inference is relatively expensive and unrealistic for real-world applications.

Several studies~\cite{rudnev2021eventhands, millerdurai2024ev2hands, jiang2024evhandpose, liu2025leveraging} tackle the 3D hand mesh reconstruction task using only event cameras during inference.
While reconstruction methods in other domains often focus on task-specific network architectures~\cite{Xiao_2025_CVPR, rudnev2023eventnerf}, event-based hand reconstruction has mainly emphasized event representations suitable for subsequent 3D hand estimation.
EventHands~\cite{rudnev2021eventhands} is a lightweight framework designed for fast hand motion reconstruction. It employs a frame-based 2D representation, termed Locally-Normalized Event Surfaces (LNES), which preserves relative temporal information and polarity within a short period of time.
Ev2Hands~\cite{millerdurai2024ev2hands} tackles the reconstruction of both hands via a point cloud-based approach. It preserves temporal information via a point cloud representation, termed Event Cloud, effectively leveraging the raw event data.
EvHandPose~\cite{jiang2024evhandpose} likewise adopts the LNES representation. In addition, it introduces motion representations based on shape flow and edges to effectively reduce motion ambiguity, and addresses the challenge of sparse annotations using a weakly supervised learning framework.
Recently, RPEP~\cite{liu2025leveraging} was proposed as a pre-training method that leverages labeled RGB images and unlabeled event data during training to improve 3D hand pose estimation.
It uses an iterative module to generate realistic pseudo-events from static images, effectively capturing non-rigid hand articulations and reducing the need for scarce event-based annotations.

These event-based 3D hand mesh reconstruction methods are restricted to fixed third-person camera views.
While there are a lot of works on 3D hand pose estimation using conventional cameras from an egocentric viewpoint~\cite{prakash2024everyday_ego, ohkawa2023AssemblyHands, mueller2017EgoRGBD, mucha2026unseendomains, Lin2022Ego2HandsPoseAD}, capturing from egocentric perception using event cameras is relatively underexplored.
Although an egocentric viewpoint offers greater flexibility and mobility than fixed third-person setups, it introduces a key challenge unique to event cameras: the wearer's motion changes the brightness across the entire scene and generates numerous background events that obscure hand-related signals.
EventEgoHands~\cite{Hara2025EventEgoHands} pioneered an egocentric approach by estimating coarse hand regions and filtering events within them to mitigate egocentric background noise.
However, it cannot distinguish between left and right hands and consistently predicts both regardless of their actual presence, which limits its practical use. 
Moreover, its reconstruction module indiscriminately applies cross-attention, hindering the effective learning of correlations between hands. 
To address these issues, we propose \textbf{EventEgoHands++}, which introduces instance-level detection and Adaptive Attention for more robust and flexible reconstruction.
%================================================================

%================================================================
\section{Proposed Method}
\label{sec:proposed_method}

We propose EventEgoHands++, a 3D hand mesh reconstruction method that relies solely on event data captured in dynamic egocentric scenes.
As illustrated in \cref{fig:model}, given an event frame $\bm{I} \in \mathbb{R}^{2 \times H \times W}$, EventEgoHands++ reconstructs 3D joint positions $\bm{J}_l, \bm{J}_r  \in \mathbb{R}^{20 \times 3}$ and mesh vertex positions $\bm{V}_l, \bm{V}_r \in \mathbb{R}^{778 \times 3}$ of both hands.
Our approach comprises two main stages: 1) the Hand Extraction Stage, which employs a Hand Detector to estimate bounding boxes and masks for the left and right hands, thereby extracting events occurring within the hand regions, and 2) the Hand Reconstruction Stage, which reconstructs the 3D hand mesh from the extracted hand events by applying Adaptive Attention.
These are described in \cref{ssec:hand_extraction_stage} and \cref{ssec:hand_reconstruction_stage}, respectively, followed by the description of the loss function in \cref{ssec:loss_functions}.
The overall workflow of EventEgoHands++ during training is summarized in \cref{alg:eventegohands_pp_training}.

\begin{algorithm}[t]
\caption{Training procedure of EventEgoHands++}
\label{alg:eventegohands_pp_training}
\begin{algorithmic}[1]
\REQUIRE
Training set $\mathcal{D} = \{(\bm{I},\, \bm{J}^{\text{gt}},\, \bm{V}^{\text{gt}},\, \bm{\theta}^{\text{gt}},\, \bm{\beta}^{\text{gt}})\}$,
pretrained Hand Detector $\mathcal{D}_{\mathrm{hand}}$,
Hand Reconstructor $\mathcal{H}_{\phi}$,
MANO model $\mathcal{M}$,
loss weights $\lambda_{\gamma}, \lambda_{\delta}, \lambda_{\epsilon}, \lambda_{\zeta}$,
learning rate $\eta$, number of epochs $E$
\ENSURE Optimized parameters $\phi$ of $\mathcal{H}_{\phi}$

\FOR{$e = 1, \ldots, E$}
    \FOR{each minibatch in $\mathcal{D}$}

        \STATE \textcolor{gray}{\textit{// Hand Extraction Stage}}
        \STATE $\bm{B}, \bm{M}, \bm{c}, \bm{s} \leftarrow \mathcal{D}_{\mathrm{hand}}(\bm{I})$
            \COMMENT{$\bm{c}$: hand-side labels (l/r), $\bm{s}$: confidence scores}
        \STATE Select $\bm{M}_{l}$ and $\bm{M}_{r}$ using hand-side labels $\bm{c}$ and confidence scores $\bm{s}$
        \STATE Determine validity indicators $m_l, m_r \in \{0,1\}$
        \STATE $\mathcal{H}_{\mathrm{vis}} \leftarrow \{h \in \{l,r\} \mid m_h = 1\}$

        \STATE \textcolor{gray}{\textit{// Hand Reconstruction Stage}}
        \IF{$\mathcal{H}_{\mathrm{vis}} = \emptyset$} \COMMENT{Condition 3: no hand detected}
            \STATE skip this sample
        \ENDIF

        \FOR{$h \in \mathcal{H}_{\mathrm{vis}}$}
            \STATE $\bm{I}_{h} \leftarrow \bm{I} \odot \bm{M}_{h}$
            \STATE $\bm{F}_{h} \leftarrow \mathrm{VisionEncoder}(\bm{I}_{h})$
        \ENDFOR

        \STATE \textcolor{gray}{\textit{// Adaptive Attention}}
        \IF{$m_l = 1$ and $m_r = 1$} \COMMENT{Condition 1: both hands detected}
            \STATE $\bar{\bm{F}}_{h} \leftarrow \mathrm{SelfAttn}_{h}(\bm{F}_{h}),\quad h \in \{l,r\}$
            \STATE $\hat{\bm{F}}_{l} \leftarrow \mathrm{CrossAttn}(\bar{\bm{F}}_{l}, \bar{\bm{F}}_{r})$
            \STATE $\hat{\bm{F}}_{r} \leftarrow \mathrm{CrossAttn}(\bar{\bm{F}}_{r}, \bar{\bm{F}}_{l})$
        \ELSE 
            \STATE \COMMENT{Condition 2: single hand detected}
            \STATE $\hat{\bm{F}}_{h} \leftarrow \mathrm{SelfAttn}_{h}(\bm{F}_{h}),\quad h \in \mathcal{H}_{\mathrm{vis}}$
        \ENDIF

        \STATE \textcolor{gray}{\textit{// MANO Decoder}}
        \FOR{$h \in \mathcal{H}_{\mathrm{vis}}$}
            \STATE $(\bm{\theta}_{h}, \bm{\beta}_{h}, \bm{t}_{h}, \bm{R}_{h})
            \leftarrow \mathrm{Decoder}(\hat{\bm{F}}_{h})$
            \STATE $\hat{\bm{J}}_{h}, \hat{\bm{V}}_{h}
            \leftarrow \mathcal{M}(\bm{\theta}_{h}, \bm{\beta}_{h}, \bm{t}_{h}, \bm{R}_{h})$
        \ENDFOR

        \STATE \textcolor{gray}{\textit{// Training Objective}}
        \STATE $\mathcal{L}_{\mathrm{hand}} \leftarrow
        \lambda_{\gamma}\mathcal{L}_{\mathrm{joints}}
        + \lambda_{\delta}\mathcal{L}_{\mathrm{interhand}}
        + \lambda_{\epsilon}\mathcal{L}_{\mathrm{vertices}}
        + \lambda_{\zeta}\mathcal{L}_{\mathrm{MANO}}$

        \STATE Update $\phi$ using AdamW with learning rate $\eta$

    \ENDFOR
\ENDFOR

\STATE \textbf{return} $\phi$
\end{algorithmic}
\end{algorithm}

\subsection{Hand Extraction Stage}
\label{ssec:hand_extraction_stage}

The Hand Detector takes an event frame $\bm{I} \in \mathbb{R}^{2 \times H \times W}$ as input and predicts instance-level hand bounding boxes 
$\bm{B} = \{\bm{b}_i\}_{i=1}^{N}$ with 
$\bm{b}_i = (x_i, y_i, w_i, h_i) \in \mathbb{R}^{4}$, where $(x_i, y_i)$ denotes the center and $(w_i, h_i)$ the width and height of the bounding box. 
All coordinates are normalized by the image width and height. 
The module also predicts a hand region mask $\bm{M} \in \mathbb{R}^{N \times H \times W}$, where $N$ denotes the number of hand instances.

We adopt an instance segmentation framework rather than a mask-only estimation approach~\cite{Hara2025EventEgoHands} because mask-only prediction tends to be unstable.
By jointly learning bounding boxes and masks, the estimation becomes more robust.
The predicted mask $\bm{M}$ is used to extract only the events within the hand region from the event frame $\bm{I}$, yielding separate masked event frames for each hand: $\bm{I}_l, \bm{I}_r \in \mathbb{R}^{2 \times H \times W}$.

We apply LNES~\cite{rudnev2021eventhands}, one of the event-frame representations, to generate $\bm{I}$ from raw events, as LNES is known to preserve temporal information through the use of temporal weighting.
We adopt YOLO26~\cite{yolo26_ultralytics}, the latest model in the widely used YOLO family for hand detection~\cite{potamias2025wilor, papo2025rohanrobusthanddetection}, as our hand detector to jointly estimate bounding boxes and segmentation masks.

\subsection{Hand Reconstruction Stage}
\label{ssec:hand_reconstruction_stage}                                                                                                                                 
              
\begin{figure}[t]
\centering
\includegraphics[width=0.8\columnwidth]{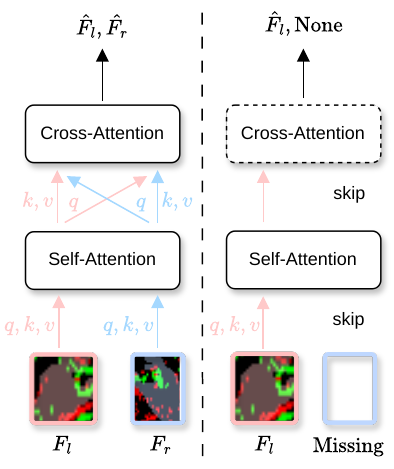}
% \vspace{-2mm}
\caption{Adaptive Attention. 
When both hands are detected, self-attention first refines intra-hand features, followed by cross-attention to capture inter-hand dependencies.
When only one hand is detected, cross-attention is skipped and only self-attention is applied.}
\label{fig:attention}
\end{figure}

The hand reconstruction stage takes filtered event frames $\bm{I}_l, \bm{I}_r \in \mathbb{R}^{2 \times H \times W}$ from the hand extraction stage and estimates the 3D joint positions $\bm{J}_l, \bm{J}_r \in \mathbb{R}^{20 \times 3}$ and mesh vertex positions $\bm{V}_l, \bm{V}_r \in \mathbb{R}^{778 \times 3}$ for each hand.

\subsubsection{Feature Extraction}
Given the masked event frames for the left and right hands, a shared backbone extracts spatial feature maps $\bm{F}_l$ and $\bm{F}_r$, each of size $C \times H' \times W'$, where $C$ denotes the channel dimension and $H',W'$ denote the spatial resolution of the feature map.
As the vision encoder, we employ EfficientNetV2-S~\cite{tan2021efficientnetv2}, which has demonstrated strong performance in various hand pose estimation architectures~\cite{ohkawa2023AssemblyHands, mucha2024inmyhands, mucha2026unseendomains}.

\subsubsection{Adaptive Attention}
To dynamically handle the presence or absence of each hand, we introduce Adaptive Attention, which selects attention operations based on the detection results.
Let $m_l, m_r \in \{0,1\}$ denote validity indicators for the left and right hands, obtained from the hand extraction stage, where $m_h=1$ indicates that the corresponding hand is detected (\eg, if $h=l$ and $m_l=1$, then it indicates that the left hand is detected).
We describe the behavior of Adaptive Attention under three hand-detection conditions: (1) both hands are detected, (2) a single hand is detected, and (3) no hands are detected.

\noindent\textbf{Condition 1: Both hands detected} ($m_l{=}1, m_r{=}1$).
Each feature map is first refined independently by self-attention to model intra-hand spatial relationships, and then the two feature maps interact through
bidirectional cross-attention to capture inter-hand dependencies:
\begin{align}
\bar{\bm{F}}_h &= \mathrm{SelfAttn}_h(\bm{F}_h), \quad h \in \{l,r\}, \label{eq:self_attn} \\
\hat{\bm{F}}_l &= \mathrm{CrossAttn}(\bar{\bm{F}}_l, \bar{\bm{F}}_r),  \label{eq:cross_attn_left} \\
\hat{\bm{F}}_r &= \mathrm{CrossAttn}(\bar{\bm{F}}_r, \bar{\bm{F}}_l). \label{eq:cross_attn_right}
\end{align}

The self-attention layer~\eqref{eq:self_attn} first refines each hand's feature map by allowing each spatial position to attend to all other positions within the same hand feature map.
In the subsequent cross-attention layers~\eqref{eq:cross_attn_left} and~\eqref{eq:cross_attn_right}, each hand's feature map serves as the query while the other hand's feature map provides the keys and values, enabling bidirectional information exchange.
This allows the model to capture inter-hand context such as relative hand positioning and coordinated finger configurations, building upon already-refined single-hand
representations.

\noindent\textbf{Condition 2: Single hand detected} ($m_l{+}m_r{=}1$).
Cross-attention is disabled since the other hand's feature map is unavailable, and only self-attention is applied to the detected hand:
\begin{equation}
\hat{\bm{F}}_h =  \mathrm{SelfAttn}_h(\bm{F}_h).
\end{equation}

\noindent\textbf{Condition 3: Neither hand detected} ($m_l{=}0, m_r{=}0$).
If neither hand is detected, the subsequent processing is skipped.

Unlike masked attention, which applies masks to attention scores inside the attention operation, our Adaptive Attention switches the attention operations themselves.
With masked attention, masking all tokens of an undetected hand corrupts the cross-attention output, which in turn destroys the features of the detected hand, while the masked attention computation itself is still executed. 
In contrast, our method skips the processing of the undetected hand entirely and processes the detected hand independently via self-attention, thereby avoiding both feature corruption and unnecessary computational overhead.
We note that the self-attention and cross-attention operations
themselves are standard, and the novelty of Adaptive Attention lies in
using per-hand visibility as an explicit control signal that determines
which operations to apply.

\subsubsection{MANO Decoder}
Finally, each refined spatial feature map is converted into a feature vector via attention pooling, which is then mapped by a linear layer to the MANO~\cite{romero2017MANO} parameter space.
These parameters are decoded by the MANO model to obtain the 3D hand joint positions $\bm{J}$ and mesh vertices $\bm{V}$.
MANO parameters for each hand consist of a pose vector $\boldsymbol{\theta} \in \mathbb{R}^{45}$, a shape vector $\boldsymbol{\beta} \in \mathbb{R}^{10}$, a translation $\boldsymbol{t} \in \mathbb{R}^3$, and a rotation $\boldsymbol{R} \in \mathbb{R}^3$.
Using the MANO model $\mathcal{M}$, we obtain sparse hand joints and dense hand mesh vertices for each hand individually as $\boldsymbol{J}, \boldsymbol{V} = \mathcal{M}(\boldsymbol{\theta}, \boldsymbol{\beta}, \boldsymbol{t}, \boldsymbol{R})$.
The resulting joint locations $\boldsymbol{J} \in \mathbb{R}^{20 \times 3}$ represent the 3D coordinates of the regressed hand joints, while the mesh vertices $\boldsymbol{V} \in \mathbb{R}^{778 \times 3}$ correspond to the 3D coordinates of the hand surface. 
For simplicity, we use the same notation for the parameters and outputs of the left and right hands unless explicitly stated otherwise.

\subsection{Training Objective}
\label{ssec:loss_functions}
To train our model, we use the loss functions adopted in the prior work~\cite{millerdurai2024ev2hands, Hara2025EventEgoHands}.

\noindent\textbf{3D Hand Joints Loss.}
The loss terms used to assess the accuracy of the estimated 3D hand joints are defined as follows.
The 3D hand joints loss $\mathcal{L_{\text{joints}}}$ is given by the L1 distance between the predicted and ground-truth joint positions, whereas the interaction hand joints loss $\mathcal{L_{\text{interhand}}}$ is the L2 distance between the predicted and ground-truth relative joint offsets of the left and right hands, measuring their positional consistency:
\begin{equation}
    \mathcal{L_{\text{joints}}} = \frac{1}{N_J}\sum_{i=1}^{N_J} \| \bm{\hat{J}_i} - \bm{J_i} \|_1,
\end{equation}

\begin{equation}
    \mathcal{L_{\text{interhand}}} = \frac{1}{N_J} \sum_{i=1}^{N_J} \| (\bm{\hat{J}}_{\text{left}, i}-\bm{\hat{J}}_{\text{right}, i})-(\bm{J}_{\text{left}, i}-\bm{J}_{\text{right}, i}) \|_2,
\end{equation}
where \( \bm{\hat{J}}_i \) and \( \bm{J}_i \) denote the predicted and ground-truth 3D coordinates of the \( i \)-th hand joint, and \( N_J \) is the total number of hand joints.

\noindent\textbf{3D Hand Mesh Vertices Loss.}
The loss terms for assessing the accuracy of the estimated 3D hand mesh vertices are defined as follows.
The 3D hand mesh vertices loss $\mathcal{L_{\text{vertices}}}$ is given by the L1 distance between the predicted and ground-truth vertex positions, encouraging accurate reconstruction of the hand mesh structure:
\begin{equation}
    \mathcal{L_{\text{vertices}}} = \frac{1}{N_V}\sum_{i=1}^{N_V} \| \bm{\hat{V}_i} - \bm{V_i} \|_1,
\end{equation}
where \( \bm{\hat{V}}_i \) and \( \boldsymbol{V}_i \) represent the predicted and ground-truth 3D coordinates of the \( i \)-th hand mesh vertex, and \( N_V \) is the total number of vertices.

\noindent\textbf{MANO Loss.} 
The MANO loss quantifies the discrepancy between the predicted and ground-truth MANO pose $\bm{\theta}$ and shape parameters $\bm{\beta}$.
This term encourages the model to produce accurate hand pose and shape representations:
\begin{equation}
    \mathcal{L_{\text{MANO}}} = \| \hat{\bm{\theta}} - \bm{\theta} \|_2 + \| \hat{\bm{\beta}} - \bm{\beta} \|_2.
\end{equation}

\noindent\textbf{Total Hand Loss.} 
These four loss terms are linearly combined to form the final training objective $\mathcal{L_{\text{hand}}}$:
\begin{align}
\mathcal{L_{\text{hand}}} = \lambda_{\gamma}\mathcal{L_{\text{joints}}} 
+ \lambda_{\delta}\mathcal{L_{\text{interhand}}}
+ \lambda_{\epsilon}\mathcal{L_{\text{vertices}}}  
+\lambda_{\zeta}\mathcal{L_{\text{MANO}}},
\end{align}
where the balancing hyperparameters $\lambda_{\gamma}, \lambda_{\delta}, \lambda_{\epsilon}$, and $\lambda_{\zeta}$ correspond to $\mathcal{L_{\text{joints}}}$, $\mathcal{L_{\text{interhand}}}$, $\mathcal{L_{\text{vertices}}}$, and $\mathcal{L_{\text{MANO}}}$, respectively.

\begin{figure}[t]
  \centering
  \includegraphics[width=\columnwidth]{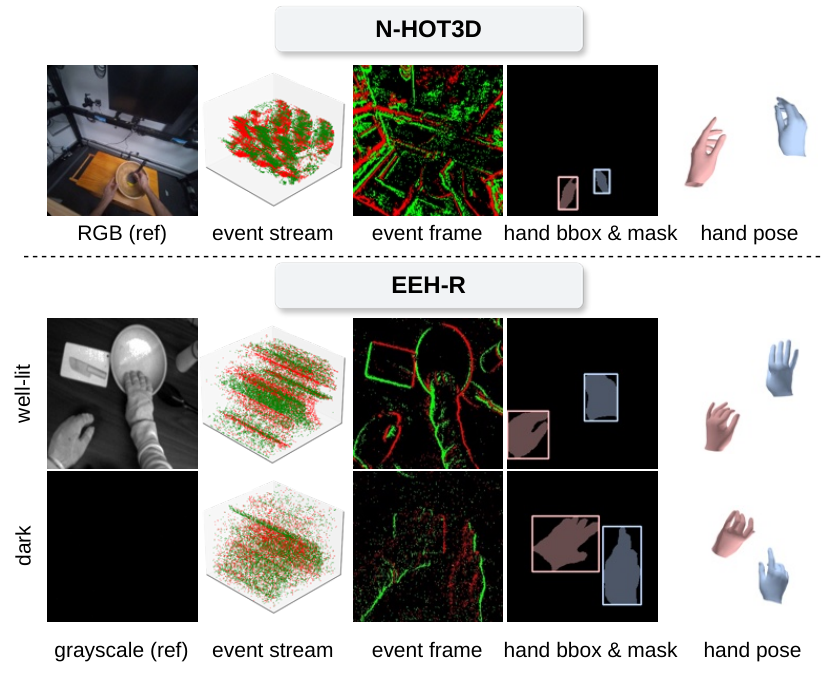}
  % \vspace{-2mm}
  \caption{Dataset samples. The top shows the synthetic dataset N-HOT3D, and the bottom shows the real-world dataset EEH-R.
}
  \label{fig:dataset_sample}
\vspace{-2mm}
\end{figure}

\section{Dataset Collection}
\label{sec:dataset_collection}

\subsection{Synthetic Dataset: N-HOT3D}
\label{ssec:synthetic_dataset}
Training our model requires event data with ground-truth annotations of both hands from an egocentric viewpoint.
However, no existing dataset provides egocentric event data with such annotations.
To address this limitation, we first construct a synthetic event dataset, N-HOT3D, by applying the event simulator v2e~\cite{hu2021v2e} to the HOT3D dataset~\cite{banerjee2025hot3d}.
Specifically, we use a subset of the Aria glasses data within HOT3D, consisting of nine subjects.
Sample data from N-HOT3D are illustrated in \cref{fig:dataset_sample} (top).

For data generation, we utilize the MANO parameters and the camera extrinsic and intrinsic parameters provided by HOT3D.
We first apply distortion correction to the videos and then convert them into event data using the simulator.
The spatial resolution of the output events is set to $346 \times 260$ pixels, matching the resolution of the DAVIS346 event camera~\cite{davis346}, which is also used in Ev2Hands~\cite{millerdurai2024ev2hands}.
Additionally, we project the provided 3D mesh ground-truth annotations onto 2D space to generate ground-truth hand segmentation masks.
We also compute bounding box annotations from the projected meshes to enable detection training.
The annotation quality of N-HOT3D depends on that of the original HOT3D annotations, since we directly use the MANO parameters and camera calibration provided by HOT3D.
The released HOT3D annotations had already been manually and visually inspected for all frames to exclude lower-quality poses.

N-HOT3D is split by both subject identity and capture sequence. The training and validation sets share the same subjects but contain disjoint capture sequences, ensuring that no identical sequence appears in both splits. 
Specifically, we use subjects P0001, P0003, P0009, P0010, P0011, P0012, and P0015 for training and validation, while the evaluation set is composed of unseen subjects, P0002 and P0014.
N-HOT3D contains a total of $480,120$ frames, divided into $334,190$ for training, $83,760$ for validation, and $62,170$ for evaluation.
Compared to the preliminary version of N-HOT3D introduced in our earlier
conference paper~\cite{Hara2025EventEgoHands}, this work extends the dataset in
several aspects.
First, we visually inspected all frames and regenerated the segmentation
masks for frames where mask generation had failed in the previous version.
As a result, the dataset composition has been updated from 447,704 frames
in the previous version to 480,120 frames in this work.
Second, we newly provide bounding box annotations computed from the
projected meshes, enabling detection training in addition to segmentation.
Finally, whereas the data split was not explicitly described in the
previous version, we clearly define the splitting protocol in this work.

\Figure[t!](topskip=0pt, botskip=0pt, midskip=0pt){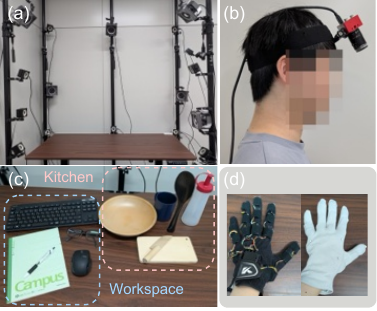}
{Capture system setup. (a) Arrangement of OptiTrack cameras, (b) head-mounted DAVIS346 event camera, (c) objects from kitchen and workspace categories, (d) MoCap gloves (left) and plain fabric gloves worn over MoCap gloves (right).
% \vspace{-2mm}
\label{fig:capture_setup}}

\subsection{Real Dataset: EEH-R}
\label{ssec:real_dataset}
To evaluate the effectiveness of our proposed method on real event camera data, we construct a real-world dataset, EEH-R. 
Our dataset includes recordings of 8 subjects performing hand-object interactions under both well-lit and dark conditions. 
In total, it provides over two hours of recording across 85 sequences with $1,019,716$ ground-truth annotations. Sample data from EEH-R are illustrated in \cref{fig:dataset_sample} (bottom).

\noindent\textbf{Capture System Setup.}
We design a capture system to simultaneously record egocentric event streams and accurate ground-truth hand/camera poses. 
An overview of our setup is illustrated in Fig.~\ref{fig:capture_setup}.
% event camera
For event data capture, we use a DAVIS346~\cite{davis346}, which provides both asynchronous events and synchronized grayscale frames. The grayscale frames are recorded at 30~fps. 

% MoCap Gloves
To obtain precise local hand poses, we use MoCap Gloves~\cite{mocap_gloves} equipped with 16 IMU sensors, capable of capturing fine-grained finger articulations including finger curvature and palm arching. 
To preserve natural hand appearance in the event data, subjects wear plain fabric gloves over the MoCap gloves, concealing the sensors while maintaining texture characteristics similar to bare hands. 
% OptiTrack
For global positioning, we use an OptiTrack system consisting of 16 cameras (6 PrimeX22 and 10 PrimeX13) managed by Motive software~\cite{motive}. 
Four optical markers are attached to both the top of the event camera, and another four optical markers are attached to the dorsal surface of each glove, enabling precise tracking of camera and hand positions.
The ground-truth hand and camera poses are recorded at 120~fps.
The average 3D calibration error of the motion-capture system was below 0.1972~mm, and the quality of the global hand and camera positions therefore depends on this calibration error.

\begin{table}[t]
\centering
\caption{EEH-R dataset statistics by scene and lighting condition.}
\label{tab:dataset_scenes}
\begin{tabular}{l l c c}
\hline
\textbf{Scene} & \textbf{Lighting} & \textbf{\#Sequences} & \textbf{\#Annotations} \\
\hline
Kitchen   & Dark  & 23 & 287,566 \\
Kitchen   & Well-lit & 19 & 238,340 \\
Workspace & Dark  & 20 & 233,470 \\
Workspace & Well-lit & 23 & 260,340 \\
\hline
\textbf{Total} &  & 85 &  1,019,716 \\
\hline
\end{tabular}
\end{table}

\noindent\textbf{Scenes.}
Our dataset covers two lighting conditions and two scene categories, recorded with eight subjects (6 male, 2 female).
For lighting conditions, we capture sequences under both well-lit (average illuminance of 457~lux) and dark (average illuminance of 3.5~lux) conditions.
For scene categories, we design kitchen and workspace scenarios to cover diverse hand-object interactions.
The kitchen scene includes everyday objects such as bowls, spoons, cups, bottles, cutting boards, and knives, while the workspace scene contains notebooks, keyboards, mice, pens, and glasses.
Each sequence lasts approximately 2 minutes, during which subjects freely interact with the objects.
Multiple sequences were recorded for each combination of lighting and scene conditions.
Detailed statistics of our dataset are summarized in Table~\ref{tab:dataset_scenes}.

\begin{figure}[t]
\centering
\includegraphics[width=\columnwidth]{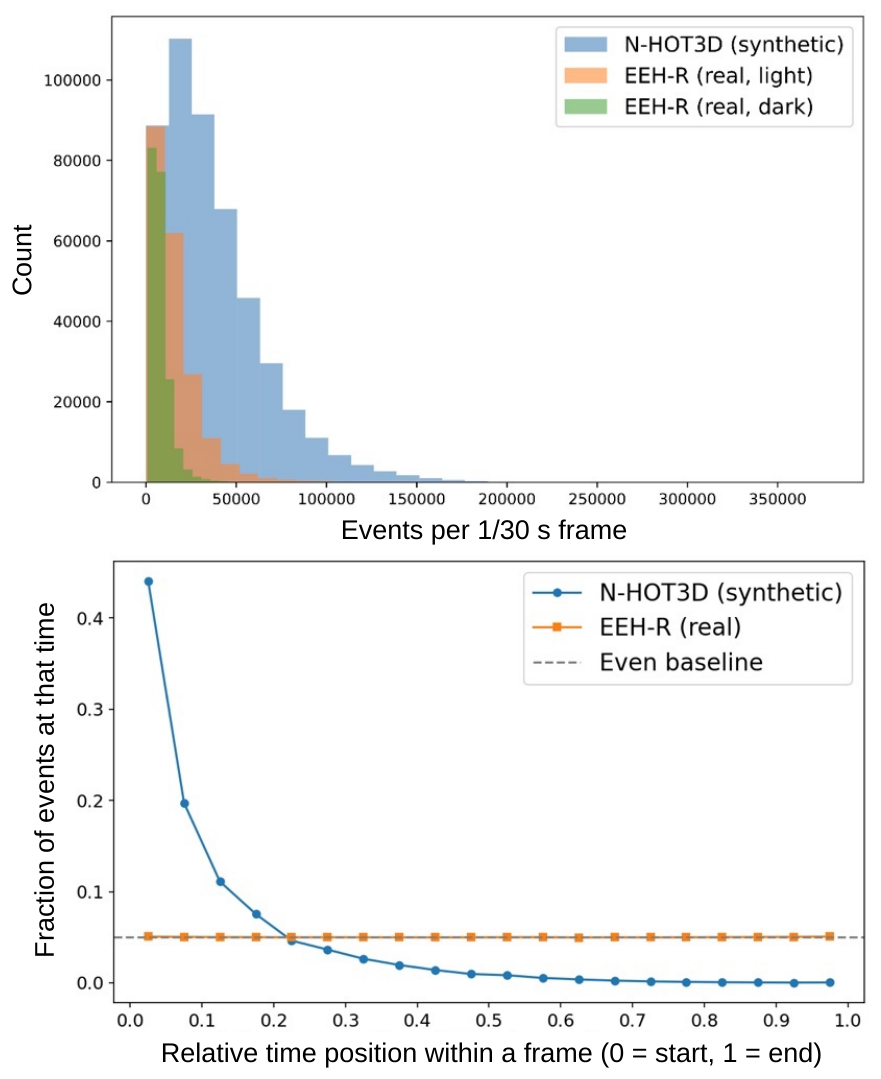}
\vspace{-4mm}
\caption{{Analysis of event characteristics in the synthetic and
real-world datasets.
(Top) Distribution of the number of events per frame (1/30\,s).
(Bottom) Temporal distribution of events within a frame  (1/30\,s).}}
\label{fig:event_analysis}
\vspace{-2mm}
\end{figure}

\noindent\textbf{Annotations.}
The MoCap gloves provide 16 joint positions per hand (wrist and three joints per finger) in the local glove coordinate system. 
Following prior hand pose dataset construction pipelines~\cite{hampali2020honnotate, brahmbhatt2020contactpose}, 
we obtain MANO parameter annotations by fitting the MANO hand model to the joint positions provided by the MoCap gloves.

In addition to 3D annotations, we construct 2D hand segmentation masks for a subset of scenes. 
For well-lit scenes, we apply SAM3~\cite{carion2025sam3segmentconcepts} to grayscale images to automatically generate ground-truth hand masks, resulting in 198,410 annotated frames.
For dark scenes, where grayscale images lack sufficient contrast, we manually annotate segmentation masks for 1,000 event frames.
As in the original HOT3D dataset, all frames in EEH-R were manually and visually inspected to remove lower-quality poses.

EEH-R is split by subject identity across training, validation, and evaluation. Specifically, subjects P05, P07, P08, and P09 are used for training, P03 and P04 for validation, and P06 and P10 for evaluation.
The EEH-R dataset contains a total of $1,019,716$ frames, divided into $636,433$ for training, $164,727$ for validation, and $218,556$ for evaluation.
Approval for EEH-R dataset recording was obtained from the ethics committee of Keio University under 2025-131.

\noindent\textbf{Domain Gap between Synthetic and Real Data.}
Compared with EEH-R, N-HOT3D involves more locomotion and head motion,
which causes intense brightness changes across the entire scene.
As shown in the top of \cref{fig:event_analysis}, N-HOT3D consequently
yields a larger number of events per frame.
The bottom of \cref{fig:event_analysis} shows how events are
distributed within the time window of each frame.
While events in the real data occur almost uniformly over time, events
in N-HOT3D are concentrated at the beginning of each frame, resulting
in an unnatural temporal distribution, since they are synthetically
generated by an event simulator.
In addition, EEH-R contains sensor noise inherent to real event
cameras.
The visual differences between the two datasets can be observed in the
event streams shown in \cref{fig:dataset_sample}.

\noindent{\textbf{Bias and Limitations.}}
EEH-R was collected in a controlled laboratory environment.
Extending the dataset to in-the-wild settings is a promising direction for future work.
Although the IMU-based sensor gloves provide accurate ground truth, they inevitably alter the visual appearance of the hands.
Adopting a marker-less motion capture system would allow bare-hand recordings, further increasing the appearance diversity of the hands.

%================================================================

\section{Experiments}
\label{sec:experiment}

\subsection{Implementation Details}
\label{ssec:implementation_details}
For training and evaluation, we follow the official train/validation/test splits of the N-HOT3D and EEH-R datasets.
Input event frames are center-cropped and resized to a resolution of $H\times W = 224 \times 224$  pixels.
All experiments were conducted using a single NVIDIA RTX 6000 Ada Generation GPU.

The Hand Detector is implemented using the YOLO26 framework~\cite{yolo26_ultralytics} along with its training and inference ecosystem.
The model is initialized from pretrained YOLO26 weights and fine-tuned for our task.
Since LNES consists of two channels, we adapt the network by introducing additional empty channels to match the expected input format.
The segmentation model is trained using Adam~\cite{kingma2015adam} with a learning rate of $1.0\times10^{-3}$ for 100 epochs with a batch size of 64.
The predicted segmentation masks are dilated using a kernel size of \(7\times7\) with one iteration to extend the hand regions.
When multiple detections of the same hand (left or right) are produced, we select the one with the highest confidence score.
The detection confidence threshold is set to 0.5 for N-HOT3D and 0.8 for EEH-R.

For the hand reconstruction stage, the EfficientNetV2-S backbone produces feature maps with channel dimension $C=1280$ and spatial resolution $H' \times W'=7 \times 7$ pixels.
The backbone is initialized with ImageNet-1K pretrained weights.
The model is trained using the AdamW~\cite{Loshchilov2017DecoupledWD} optimizer with a learning rate of $2.0\times10^{-5}$ for 30 epochs with a batch size of 32.
The total hand loss weights are set as $\lambda_{\gamma}=2.0$, $\lambda_{\delta}=1.0$, $\lambda_{\epsilon}=2.0$, and $\lambda_{\zeta}=1.0$.

% The weights of MANO parameter fitting are set as
% $\lambda_{\text{bone}}=0.1$,
% $\lambda_{\text{angle}}=0.01$,
% $\lambda_{\text{smooth}}=0.1$,
% $\lambda_{\beta}=0.1$,
% and $\lambda_{\theta}=1\times10^{-5}$.

\subsection{Baseline Methods}
We compare our method with the following baselines to show its performance in event-based 3D hand mesh reconstruction.

\label{ssec:baseline_methods}
\noindent\textbf{EventHands}~\cite{rudnev2021eventhands} is a frame-based approach that operates solely on event data. 
As it is regarded as one of the state-of-the-art methods for third-person event-based hand reconstruction, we adopt it as a baseline.
Since this method predicts only a single hand at a time, we trained two separate models for the left and right hands on N-HOT3D and EEH-R.

\noindent\textbf{Ev2Hands}~\cite{millerdurai2024ev2hands} leverages only event data and extracts the features using a point cloud-based representation.
% is a point cloud-based approach that also uses only event data.
It is the only prior method that predicts both hands simultaneously.
Since it requires event data annotated as left hand, right hand, or background, whose labels are not available in N-HOT3D or EEH-R, we first train the model on the annotated Ev2Hands-S~\cite{millerdurai2024ev2hands} dataset and then fine-tune it on N-HOT3D and EEH-R.

\noindent\textbf{EventEgoHands}~\cite{Hara2025EventEgoHands} is the first egocentric hand mesh reconstruction method.
This method estimates hand regions using mask-only estimation without distinguishing between left and right hands.
The reconstruction stage is point cloud-based and employs only cross-attention between hand features to estimate hand meshes.
Since this method does not differentiate between left and right hands, it always outputs two hands regardless of their presence.

\begin{figure*}[t]
  \centering
  \includegraphics[width=\textwidth]{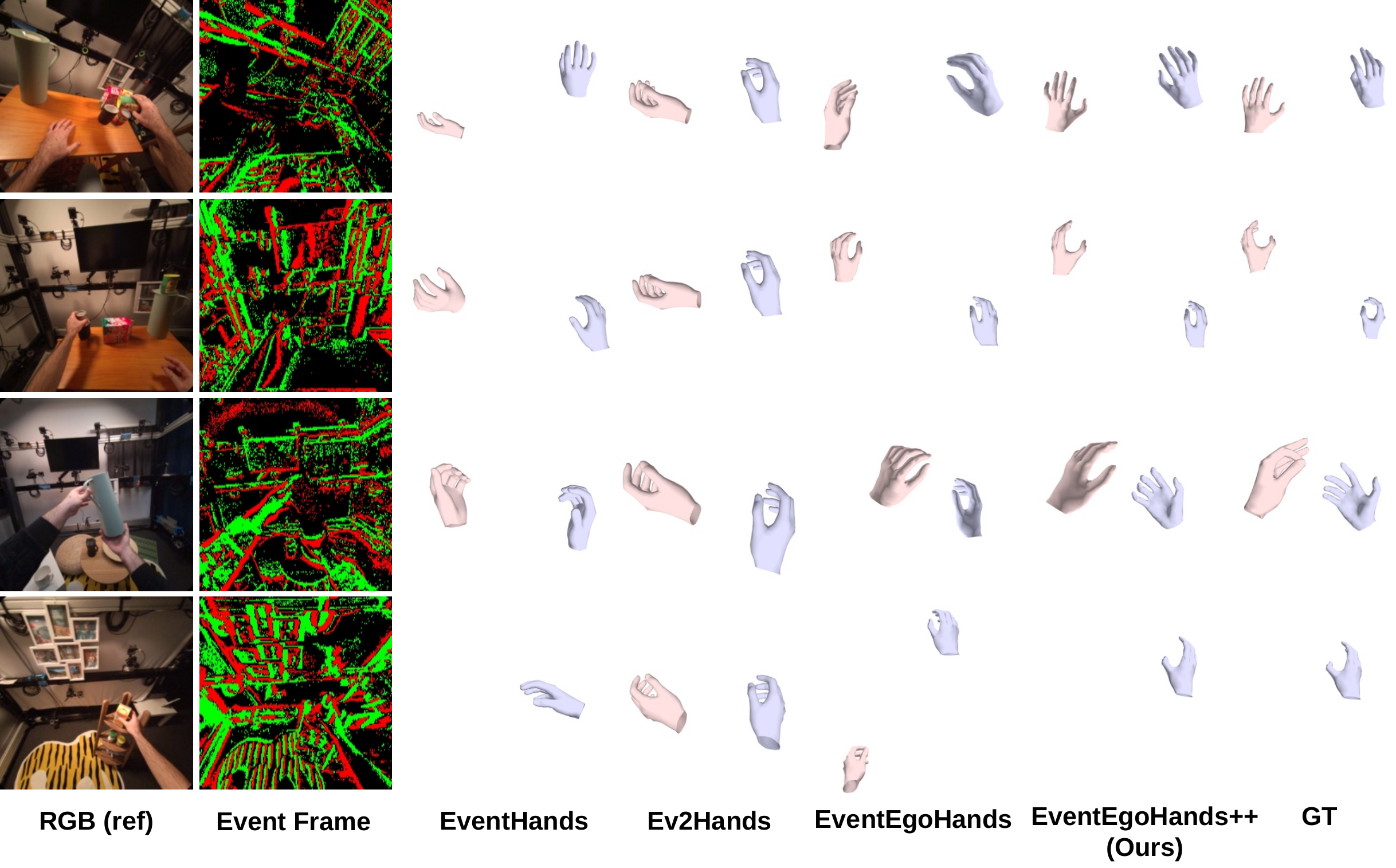}
  \vspace{-4mm}
  \caption{Qualitative evaluation on N-HOT3D. We compared our method with EventHands~\cite{rudnev2021eventhands}, Ev2Hands~\cite{millerdurai2024ev2hands} and EventEgoHands~\cite{Hara2025EventEgoHands}. RGB images were not used as input and are shown for reference only.
  In order from the top, the scenes show ``pouring from a can'', ``picking up a can'', ``holding a pot'', and ``picking up a carton''.
  }
  \label{fig:result}
  \vspace{-2mm}
\end{figure*}

\subsection{Evaluation Metrics}
\label{ssec:evaluation_metrics}

Following Ev2Hands~\cite{millerdurai2024ev2hands}, we adopt the Percentage of Correct Keypoints (PCK) and the area under the PCK curve (AUC) with thresholds ranging from 0 to 100 millimeters (mm).
We also use the Relative AUC (R-AUC) and Right-root Relative AUC (RR-AUC) to assess the 3D hand pose estimation performance. 
The R-AUC is computed from joint positions expressed relative to the wrist of each hand, thereby measuring the accuracy of 3D joint positions for each hand independently.
In contrast, RR-AUC is computed by expressing the joint positions of both hands relative to the right wrist, thereby evaluating the relative 3D configuration between the two hands.
In addition, we report the Mean Per Joint Position Error (MPJPE) and Mean Per Vertex Position Error (MPVPE) in millimeters, which are standard metrics for 3D hand mesh reconstruction. Both MPJPE and MPVPE are computed after aligning each hand by its wrist joint position.
For all the metrics above, only the hands successfully detected by the Hand Detector are included in the evaluation, and undetected hands are excluded from the aggregation of joint and mesh errors.
The detection performance itself is evaluated separately using the metrics described below.

Furthermore, to evaluate the performance of the Hand Detector, we employ the mean Average Precision (mAP) at different Intersection over Union (IoU) thresholds, namely mAP@50 and mAP@50--95. 
mAP@50 denotes the mAP at an IoU threshold of 0.5, while mAP@50--95 is averaged over IoU thresholds from 0.5 to 0.95 with a step size of 0.05. 
These metrics are widely used in instance segmentation and detection tasks, including the YOLO series~\cite{yolo26_ultralytics}.

%================================================================

\begin{table}[t]
    \centering
    \caption{\textbf{Quantitative evaluation on N-HOT3D.} We report the mean $\pm$ standard deviation over three runs with different random seeds. The best values are shown in \textbf{bold}.}
    \label{table:quant}
     \setlength{\tabcolsep}{3pt}
     \scalebox{0.78}{
    \begin{tabular}{lcccc}
        \toprule
          & R-AUC ($\uparrow$) &  RR-AUC ($\uparrow$) & MPJPE [mm] ($\downarrow$) & MPVPE [mm] ($\downarrow$) \\
        \midrule
        EventHands~\cite{rudnev2021eventhands}  & 0.253 $\pm$ 0.009 & 0.190 $\pm$ 0.008 & 105.62 $\pm$ 1.40 & 97.29 $\pm$ 1.29 \\
        Ev2Hands~\cite{millerdurai2024ev2hands} & 0.236 $\pm$ 0.001 & 0.209 $\pm$ 0.000 & 112.56 $\pm$ 0.46 & $107.00 \pm  0.44$ \\
        EventEgoHands~\cite{Hara2025EventEgoHands} & 0.417 $\pm$ 0.004 & 0.261 $\pm$ 0.021 & 64.83 $\pm$ 0.56 & 60.53 $\pm$ 0.52 \\
        \rowcolor{gray11}
        Ours & \textbf{0.661 $\pm$ 0.004} & \textbf{0.528 $\pm$ 0.003} & \textbf{43.01 $\pm$ 0.27} & $\mathbf{39.96 \pm 0.26}$  \\
        \bottomrule
    \end{tabular}
    }
\end{table}

\section{Results}
\label{sec:results}

\begin{table*}[t]
    \centering
    \caption{\textbf{Quantitative evaluation on EEH-R under different lighting conditions.} All values are reported as mean $\pm$ standard deviation over three runs with different random seeds. The best values are shown in \textbf{bold}.}
    \label{table:realdata_lighting}
    \begin{tabular}{llcccc}
        \toprule
        Method & Lighting & R-AUC ($\uparrow$) & RR-AUC ($\uparrow$) & MPJPE [mm] ($\downarrow$) & MPVPE [mm] ($\downarrow$) \\
        \midrule
        \multirow{3}{*}{EventHands~\cite{rudnev2021eventhands}}
            & All      & $0.583\pm0.004$ & $0.287\pm0.004$ & $42.08\pm0.44$ & $39.29\pm0.41$ \\
            & Well-lit & $0.591\pm0.005$ & $0.298\pm0.006$ & $41.38\pm0.59$ & $38.56\pm0.53$ \\
            & Dark     & $0.576\pm0.003$ & $0.278\pm0.004$ & $42.71\pm0.35$ & $39.95\pm0.34$ \\
        \midrule
        \multirow{3}{*}{Ev2Hands~\cite{millerdurai2024ev2hands}}
            & All      & $0.502\pm0.002$ & $0.250\pm0.004$ & $52.02\pm0.54$ & $48.59\pm0.48$ \\
            & Well-lit & $0.492\pm0.004$ & $0.262\pm0.002$ & $53.57\pm0.87$ & $50.05\pm0.80$ \\
            & Dark     & $0.511\pm0.005$ & $0.239\pm0.006$ & $50.64\pm0.69$ & $47.29\pm0.61$ \\
        \midrule
        \multirow{3}{*}{EventEgoHands~\cite{Hara2025EventEgoHands}}
            & All      & $0.478\pm0.008$ & $0.324\pm0.012$ & $56.59\pm1.69$ & $52.94\pm1.64$ \\
            & Well-lit & $0.481\pm0.003$ & $0.340\pm0.005$ & $56.27\pm1.13$ & $52.65\pm1.05$ \\
            & Dark     & $0.475\pm0.018$ & $0.310\pm0.022$ & $56.95\pm3.35$ & $53.27\pm3.12$ \\
        \midrule
        \rowcolor{gray11}
        % \multirow{3}{*}{Ours}
            & All      & ${0.691\pm0.003}$ & ${0.551\pm0.005}$ & ${34.18\pm0.22}$ & ${32.16\pm0.23}$ \\
        \rowcolor{gray11}
           Ours & Well-lit & $\mathbf{0.693\pm0.003}$ & $\mathbf{0.606\pm0.006}$ & $\mathbf{32.68\pm0.30}$ & $\mathbf{30.68\pm0.29}$ \\
        \rowcolor{gray11}
            & Dark     & ${0.689\pm0.003}$ & ${0.476\pm0.005}$ & ${35.57\pm0.18}$ & ${33.53\pm0.22}$ \\
        \bottomrule
    \end{tabular}
\end{table*}

\begin{figure*}[t]
  \centering
  \includegraphics[width=\textwidth]{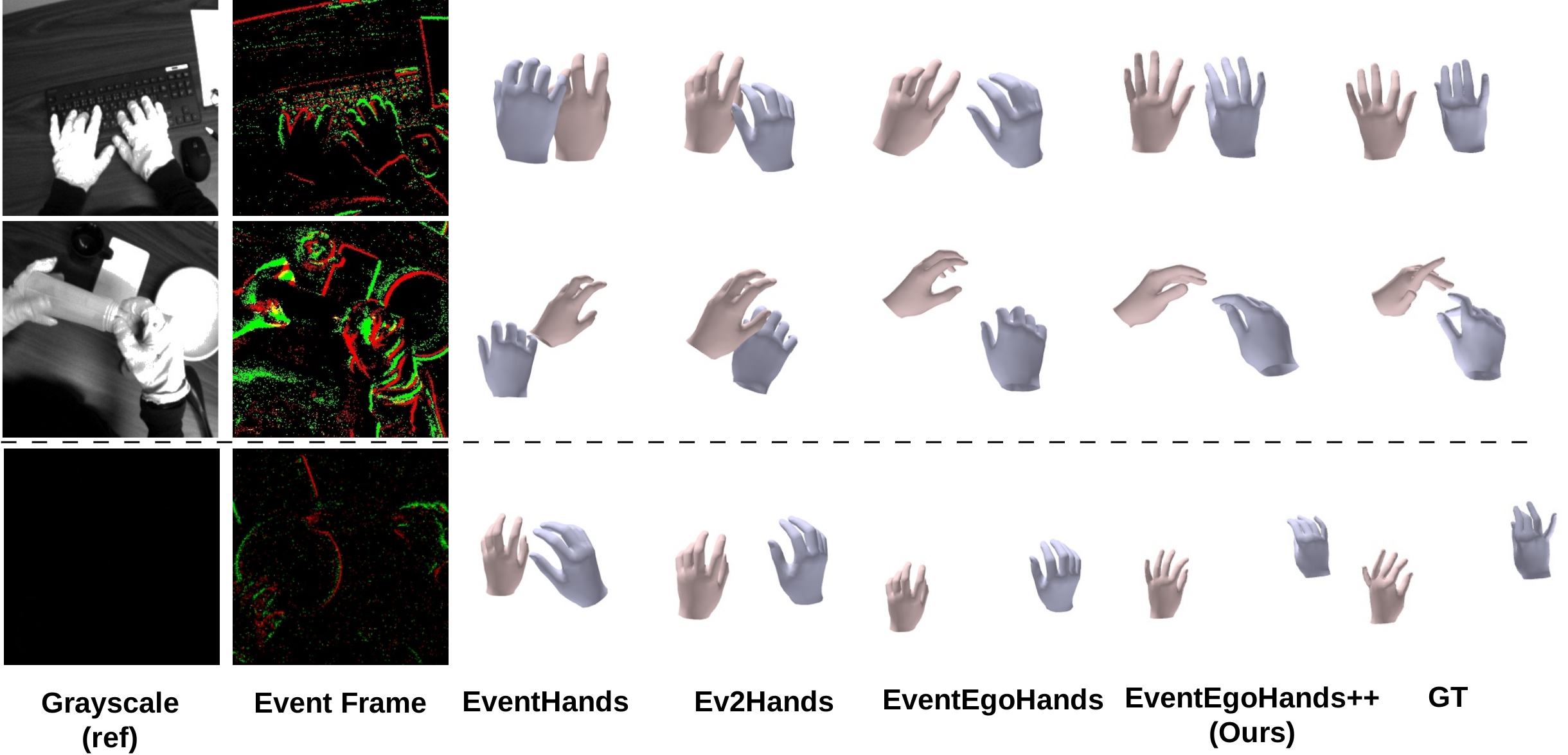}
  \caption{Qualitative evaluation on EEH-R. We compared our method with EventHands~\cite{rudnev2021eventhands}, Ev2Hands~\cite{millerdurai2024ev2hands} and EventEgoHands~\cite{Hara2025EventEgoHands}. Grayscale images were not used as input and are shown for reference only. The grayscale images at the bottom are actual captured images, which appear almost empty due to the low-light environment.
  In order from the top, the scenes show ``typing on a keyboard'', ``holding a bottle'', and ``using a spoon''.
  }
  \label{fig:result_real}
  \vspace{-4mm}
\end{figure*}

\subsection{Evaluation on N-HOT3D dataset}
\label{ssec:nhot3d}

\noindent\textbf{Quantitative Evaluation.}
\cref{table:quant} presents the quantitative results on the N-HOT3D dataset.
Our method significantly outperforms all existing methods across all metrics.
Compared to the closest baseline EventEgoHands, our enhanced model demonstrates superior precision, reducing MPJPE and MPVPE by 33.7\% and 34.0\%, respectively.

\noindent\textbf{Qualitative Evaluation.}
\cref{fig:result} shows qualitative comparisons on the N-HOT3D dataset.
Our method reconstructs hand meshes that are closest to the ground truth compared with other methods.
EventHands, which estimates each hand independently, often produces inconsistent hand positions and fails to capture the spatial relationship between the two hands.
Ev2Hands, which is based on point cloud processing, limits the number of input event points due to computational constraints. 
In egocentric settings where a large number of events are generated, uniform subsampling reduces the number of events belonging to the hand region, which prevents the model from  learning sufficiently detailed hand shape representations and often results in nearly identical hand poses.
EventEgoHands improves the spatial consistency between the two hands by filtering hand regions.
However, the reconstructed hand poses lack diversity and tend to produce similar poses across different scenes.

In contrast, the proposed method reconstructs more accurate hand shapes and better represents both inter-hand relationships and detailed finger motions such as fingertip positions and hand rotations.
Notably, in scenes where only a single hand is visible, our approach predicts an accurate pose for the visible hand while correctly identifying the absence of the other. 
Whereas existing methods often produce incorrect poses or erroneously predict a second hand that is not present, our framework remains robust to such scenarios.
This improvement is attributed to the Hand Detector and Adaptive Attention, which effectively learn both single-hand features and interactions between the two hands.

\subsection{Evaluation on EEH-R dataset}
\label{ssec:eehr}

\noindent\textbf{Quantitative Evaluation.}
\cref{table:realdata_lighting} shows the quantitative results on the real-world EEH-R dataset.
Our method again achieves the best performance across all evaluation metrics.
Compared with the best existing method, our method reduces MPJPE by 7.90\,mm (18.8\%) and MPVPE by 7.13\,mm (18.1\%).
The improvement margin is smaller than that observed on the synthetic dataset.
We attribute this difference to the characteristics of real-world data, where the camera is typically closer to the hands, resulting in larger hand regions and relatively easier detection.
In contrast, the synthetic N-HOT3D dataset contains scenarios where the hands appear smaller due to larger interaction distances and fisheye distortion correction, making hand detection and reconstruction more challenging.

Another notable observation is that the third-person baselines, EventHands and Ev2Hands, outperform the egocentric prior work EventEgoHands on several metrics. We attribute this to the characteristics of the real-world data. 
All sequences in EEH-R are captured during desk-based tasks, where the camera-to-hand distance is short and nearly constant, and the hands occupy a large portion of the frame, making hand detection itself easier, as also indicated by the mAP in \cref{tab:mask_results}. 
Furthermore, as shown in \cref{fig:event_analysis}, the number of events per frame in EEH-R is smaller than in N-HOT3D, so hand-related events are relatively dominant within each frame, creating a setting closer to the third-person scenario where only hand events are observed.
In addition, R-AUC, MPJPE, and MPVPE evaluate hand pose errors in wrist-relative coordinates, and the local hand poses show little difference across the baselines, as seen in \cref{fig:result_real}. 
In contrast, EventEgoHands surpasses the third-person baselines on RR-AUC, which evaluates the relative position between the two hands, and this advantage is also evident in \cref{fig:result_real}.

\noindent\textbf{Qualitative Evaluation.}
\cref{fig:result_real} shows qualitative results on the EEH-R dataset.
Consistent with the quantitative results, the differences between methods are smaller than those observed on the synthetic dataset.
This is because EEH-R consists of static desk-based tasks with less subject motion than N-HOT3D.

Nevertheless, existing methods still exhibit limitations similar to those observed on the synthetic dataset.
They often fail to accurately capture inter-hand relationships and fine details of intra-hand shape.
In contrast, our method reconstructs hand meshes that are closest to the ground truth and produces more diverse hand poses across different scenes.
These results demonstrate that the proposed method remains effective when applied to real event camera data, including challenging low-light environments.

\begin{table}[t]
\centering
\caption{Comparison of inference speed (FPS) across methods. Values are reported as the mean $\pm$ standard deviation.}
\label{tab:fps_comparison}
\setlength{\tabcolsep}{3pt}
\scalebox{0.9}{
\begin{tabular}{lccc}
\toprule
Method & Segmentation & Hand & Overall \\
\midrule
EventHands~\cite{rudnev2021eventhands}        & --              & $415.88\pm143.65$ & $415.88\pm143.65$ \\
Ev2Hands~\cite{millerdurai2024ev2hands}       & --              & $11.16\pm0.96$    & $11.16\pm0.96$    \\
EventEgoHands~\cite{Hara2025EventEgoHands}    & $209.44\pm39.24$ & $12.53\pm0.52$    & $12.41\pm0.35$    \\
\midrule
Ours (Single Hand) & $149.94\pm36.18$ & $62.46\pm16.27$ & $45.42\pm9.10$ \\
Ours (Both Hands)  & $160.28\pm35.91$ & $64.09\pm12.80$ & $39.87\pm6.63$ \\
\bottomrule
\end{tabular}
}
\end{table}

\noindent{\textbf{Runtime Analysis.}}
\cref{tab:fps_comparison} shows the inference speed of the proposed method and  the baselines in frames per second (FPS), reported as the mean $\pm$ standard deviation over 100 samples.
EventHands is extremely fast. However, as shown in \cref{table:quant} and \cref{table:realdata_lighting}, its accuracy is insufficient.
Ev2Hands and EventEgoHands rely on point-cloud-based processing, which involves higher-dimensional inputs than image-based processing, resulting in low FPS, particularly for hand mesh reconstruction.
In contrast, the proposed method is fast in both segmentation and hand mesh reconstruction, achieving practical inference speed in the overall pipeline.
Moreover, when only one hand is detected, inference is slightly faster because certain operations, such as cross-attention in Adaptive Attention, are skipped.

% \begin{table}[t]
%     \centering
%     \caption{Performance comparison of segmentation methods on synthetic and real datasets.}
%     \label{tab:mask_results}
%     \scalebox{0.9}{
%         \begin{tabular}{l l cc}
%         \hline
%         Dataset & Method & mAP@50 ($\uparrow$) & mAP@50--95 ($\uparrow$) \\
%         \hline
%         \multirow{2}{*}{N-HOT3D} 
%         & U-Net in EventEgoHands & 0.312 & 0.051 \\
%         & Hand Detector & \textbf{0.866} & \textbf{0.449} \\
%         \hline
%         \multirow{2}{*}{EEH-R} 
%         & U-Net in EventEgoHands & 0.930 & 0.657 \\
%         & Hand Detector & \textbf{0.938} & \textbf{0.686} \\
%         \hline
%         \end{tabular}
%     }
%     \vspace{-4mm}
% \end{table}

\begin{table*}[t]
    \centering
    \caption{Performance comparison of segmentation methods on synthetic and real datasets. All values are reported as mean $\pm$ standard deviation over three runs with different random seeds.
    The best values are shown in \textbf{bold}.}
    \label{tab:mask_results}
    \resizebox{\textwidth}{!}{
    \begin{tabular}{l l cc cc cc}
        \toprule
        & & \multicolumn{2}{c}{All} 
        & \multicolumn{2}{c}{Well-lit} 
        & \multicolumn{2}{c}{Dark} \\
        \cmidrule(lr){3-4}
        \cmidrule(lr){5-6}
        \cmidrule(lr){7-8}
        Dataset & Method
        & mAP@50 ($\uparrow$) 
        & mAP@50--95 ($\uparrow$)
        & mAP@50 ($\uparrow$) 
        & mAP@50--95 ($\uparrow$)
        & mAP@50 ($\uparrow$) 
        & mAP@50--95 ($\uparrow$ )\\
        \midrule

        \multirow{2}{*}{N-HOT3D}
        & U-Net in EventEgoHands
        & 0.272 $\pm$ 0.025
        & 0.042 $\pm$ 0.006
        & -- 
        & -- 
        & -- 
        & -- \\
        & Hand Detector
        & \textbf{0.750 $\pm$ 0.013}
        & \textbf{0.407 $\pm$ 0.012}
        & -- 
        & -- 
        & -- 
        & -- \\

        \midrule

        \multirow{2}{*}{EEH-R}
        & U-Net in EventEgoHands
        & \textbf{0.928 $\pm$ 0.003}
        & 0.655 $\pm$ 0.002
        & \textbf{0.929 $\pm$ 0.003}
        & 0.656 $\pm$ 0.002
        & \textbf{0.899 $\pm$ 0.025}
        & 0.527 $\pm$ 0.021 \\
        & Hand Detector
        & 0.901 $\pm$ 0.003
        & \textbf{0.657 $\pm$ 0.005}
        & 0.901 $\pm$ 0.003
        & \textbf{0.657 $\pm$ 0.005}
        & 0.887 $\pm$ 0.095
        & \textbf{0.639 $\pm$ 0.091} \\

        \bottomrule
    \end{tabular}
    }
\end{table*}

\begin{figure}[t]
  \centering
  \includegraphics[width=\columnwidth]{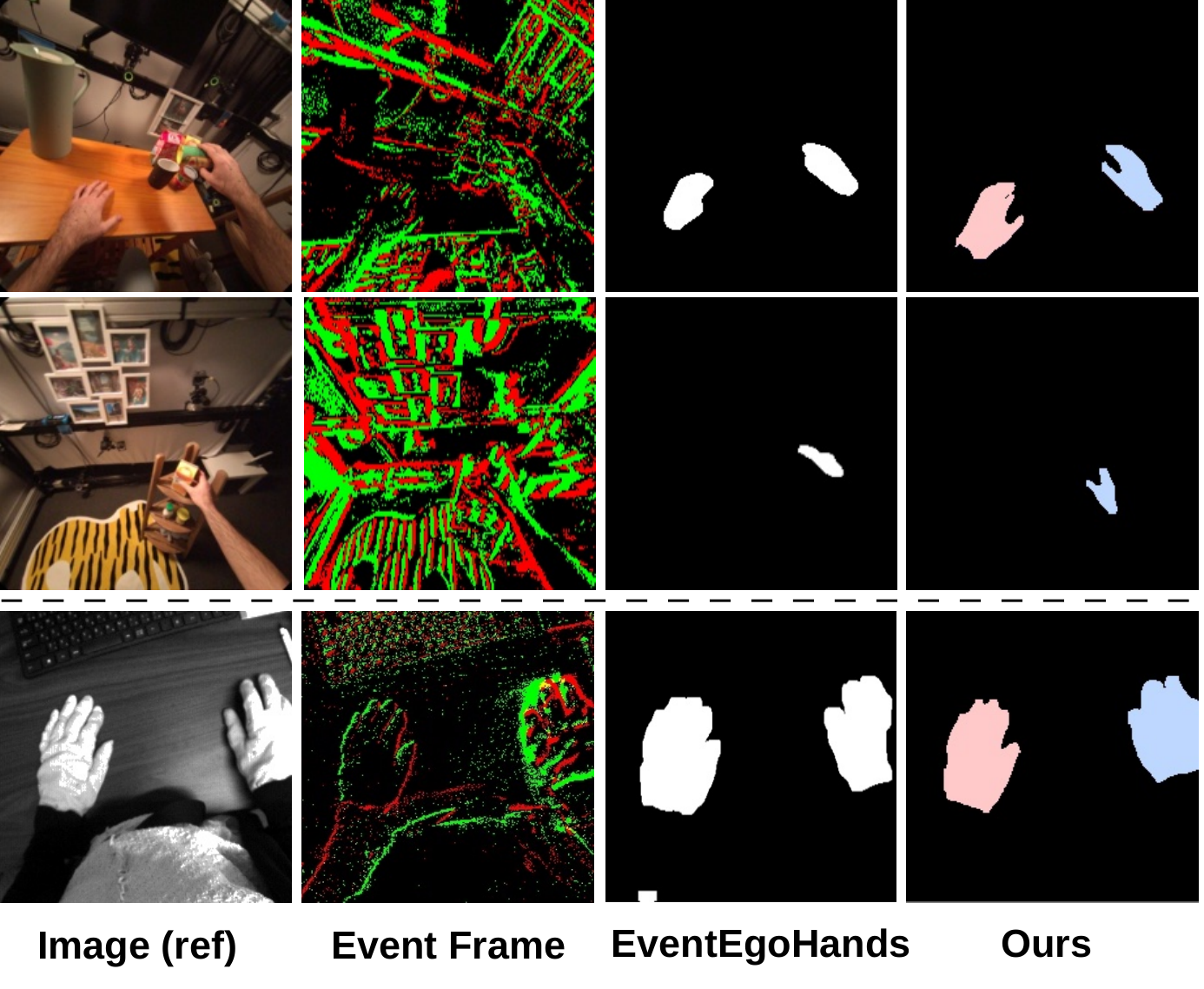}
  \vspace{-4mm}
  \caption{Hand segmentation results. The top two rows correspond to N-HOT3D and the bottom row corresponds to EEH-R. Since the segmentation produced by EventEgoHands is a binary classification, the mask region is shown in white, whereas our method uses different colors for the left and right hands.}
  \label{fig:mask_result}
\end{figure}

\subsection{Hand Segmentation Performance}
\label{ssec:hand_segmentation_performance}
We conduct an experiment to validate the effectiveness of the proposed Hand Detector.
\cref{tab:mask_results} compares the performance of the mask segmentation methods on the N-HOT3D and EEH-R datasets. 
Qualitative segmentation results are shown in \cref{fig:mask_result}.

The Hand Detector significantly outperforms the U-Net on the N-HOT3D dataset, improving mAP@50 from 0.272 to 0.750 and mAP@50--95 from 0.042 to 0.407.
The improved accuracy on N-HOT3D, where hands occupy only a small portion of the frame, suggests the effectiveness of instance segmentation that jointly learns bounding boxes and masks, compared to the simple mask estimation performed by U-Net.

On EEH-R, we evaluate the performance under each lighting condition.
Since EEH-R contains fewer events per frame and the hands are more visible within the scene, the mAP values are higher than those on N-HOT3D, and the Hand Detector performs slightly worse than the U-Net in terms of mAP@50.
However, the Hand Detector outperforms the U-Net on mAP@50--95, which evaluates mask overlap more strictly.
In particular, the improvement is substantial under the dark condition.

\begin{table}[t]
    \centering
    \caption{\textbf{Ablation study on the proposed components on N-HOT3D.} All values are reported as mean $\pm$ standard deviation over three runs with different random seeds. The best values are shown in \textbf{bold}.}
    \label{table:ablation}
     \setlength{\tabcolsep}{3pt}
     \scalebox{0.75}{
    \begin{tabular}{cc|cccc}
        \toprule
        Detector & Attention & R-AUC ($\uparrow$) & RR-AUC ($\uparrow$) & MPJPE [mm] ($\downarrow$) & MPVPE [mm] ($\downarrow$) \\
        \midrule
         &  & 0.548 $\pm$ 0.002 & 0.389 $\pm$ 0.002 & 46.79 $\pm$ 0.21 & 43.79 $\pm$ 0.19 \\
         & \checkmark & 0.553 $\pm$ 0.002 & 0.394 $\pm$ 0.002 & 47.15 $\pm$ 0.19 & 43.78 $\pm$ 0.18 \\
        \checkmark &  & 0.652 $\pm$ 0.001 & 0.469 $\pm$ 0.002 & 44.01 $\pm$ 0.03 & 41.04 $\pm$ 0.03 \\
        \checkmark & \checkmark & \textbf{0.661 $\pm$ 0.004} & \textbf{0.528 $\pm$ 0.003} & \textbf{43.01 $\pm$ 0.27} & \textbf{39.96 $\pm$ 0.26} \\
        \bottomrule
    \end{tabular}}
\end{table}

\begin{table}[t]
    \centering
    \caption{\textbf{Effect of introducing the Hand Detector to existing methods on N-HOT3D.} 
    All values are reported as mean $\pm$ standard deviation over three runs with different random seeds.
    \underline{Underline} indicates improvement within each method.}
    \label{table:seg_comparison}
     \setlength{\tabcolsep}{3pt}
     \scalebox{0.7}{
    \begin{tabular}{l|cccc}
        \toprule
        Method & R-AUC ($\uparrow$) & RR-AUC ($\uparrow$) & MPJPE [mm] ($\downarrow$) & MPVPE [mm] ($\downarrow$) \\
        \midrule
        EventHands~\cite{rudnev2021eventhands}  & 0.253 $\pm$ 0.009 & 0.190 $\pm$ 0.008 & 105.62 $\pm$ 1.40 & 97.29 $\pm$ 1.29 \\
        EventHands w/ Detector  & \underline{0.293 $\pm$ 0.001}  & \underline{0.234 $\pm$ 0.007} &  108.09 $\pm$ 4.77 & 99.33 $\pm$ 4.31 \\
        \midrule
        Ev2Hands~\cite{millerdurai2024ev2hands} & 0.236 $\pm$ 0.001 & 0.209 $\pm$ 0.000 & 112.56 $\pm$ 0.46 & 107.00 $\pm$  0.44 \\
        Ev2Hands w/ Detector  & \underline{0.242 $\pm$ 0.001} & \underline{0.214 $\pm$ 0.001} & \underline{110.05 $\pm$ 0.07} & \underline{104.59 $\pm$ 0.06} \\
        \midrule
        EventEgoHands~\cite{Hara2025EventEgoHands} & 0.417 $\pm$ 0.004 & 0.261 $\pm$ 0.021 & 64.83 $\pm$ 0.56 & 60.53 $\pm$ 0.52 \\
        EventEgoHands  w/ Detector & \underline{0.437 $\pm$ 0.013} & \underline{0.277 $\pm$ 0.012} & 68.64 $\pm$ 4.65 & 63.86 $\pm$ 4.33 \\
        
        \bottomrule
    \end{tabular}}
\end{table}

\subsection{Ablation Study}
\label{ssec:ablation}

We conduct ablation studies to evaluate the contribution of each component to the proposed framework.
All experiments are performed on the N-HOT3D dataset.

\noindent\textbf{Effect of the proposed components.}
\cref{table:ablation} shows the results when removing the Hand  Detector and the Adaptive Attention.
Even when both components are removed, the proposed framework still outperforms existing methods.
This suggests that the frame-based architecture, which jointly models two hands, is more effective than previous single-hand or point-cloud-based approaches.
Note that without the Hand Detector, the two hands cannot be localized
or separated, so both hands are always treated as present
($m_l{=}m_r{=}1$) and the Adaptive Attention always applies both
self-attention and cross-attention.

When Adaptive Attention alone is introduced without the Hand Detector, the R-AUC slightly improves, while MPJPE slightly degrades.
This indicates that attention alone is insufficient when hand regions are not properly localized; without the detector, the attention mechanism is always forced to process the entire frame, including background noise, rather than focusing on the actual hand instances.

In contrast, introducing the Hand Detector improves
performance across all metrics, and combining it with Adaptive
Attention yields further gains.
In particular, the gain from Adaptive Attention is most
pronounced in RR-AUC, which evaluates the relative position between
the two hands, improving from 0.469 to 0.528 (a 12.6\% relative
improvement).
These results indicate a complementary division of roles: the Hand
Detector is the primary source of overall accuracy by localizing and
identifying each hand, whereas Adaptive Attention specifically
strengthens the inter-hand relative positioning.

\noindent\textbf{General applicability of the Hand  Detector.}
\cref{table:seg_comparison} evaluates the effect of integrating the Hand Detector into existing methods.
Across all three methods, R-AUC and RR-AUC consistently improve after incorporating the proposed Hand Detector.
These results indicate that extracting hand regions from event data is generally beneficial and can improve a wide range of existing hand reconstruction methods.

\noindent{\textbf{Loss Functions.}}
\cref{table:ablation_loss} shows the ablation study on the loss components.
We adopt the loss functions used in Ev2Hands and EventEgoHands, which also output two hands.
The results show that combining all loss components achieves the best performance.

\begin{table}[t]
    \centering
    \caption{\textbf{Ablation study of loss components on N-HOT3D.}
    The best values are shown in \textbf{bold}.}
    \label{table:ablation_loss}
    \setlength{\tabcolsep}{3pt}
    \scalebox{0.70}{
    \begin{tabular}{cccccccc}
        \toprule
        $\mathcal{L}_{\text{MANO}}$
        & $\mathcal{L}_{\text{interhand}}$
        & $\mathcal{L}_{\text{joints}}$
        & $\mathcal{L}_{\text{vertices}}$
        & R-AUC ($\uparrow$)
        & RR-AUC ($\uparrow$)
        & MPJPE [mm] ($\downarrow$)
        & MPVPE [mm] ($\downarrow$) \\
        \midrule
        \checkmark &            &            &            & 0.095 & 0.073 & 153.64 & 142.10 \\
        \checkmark & \checkmark &            &            & 0.311 & 0.239 & 80.12 & 74.00 \\
        \checkmark & \checkmark & \checkmark &            & 0.657 & 0.521 & 43.50 & 40.43 \\
        \checkmark & \checkmark & \checkmark & \checkmark & \textbf{0.661} & \textbf{0.528} &  \textbf{43.01} & \textbf{39.96} \\
        \bottomrule
    \end{tabular}
    }
\end{table}

\begin{table}[t]
    \centering
    \caption{\textbf{Comparison of Vision Encoders on N-HOT3D.} The best values are shown in \textbf{bold}.}
    \label{table:vision_encoder}
    \setlength{\tabcolsep}{3pt}
    \scalebox{0.75}{
    \begin{tabular}{l|c|cccc}
        \toprule
        Vision Encoder & Params [M] & R-AUC ($\uparrow$)& RR-AUC ($\uparrow$) & MPJPE ($\downarrow$) & MPVPE ($\downarrow$) \\
        \midrule
        ResNet50~\cite{he2016resnet} & 23.5 & 0.653 & 0.519 & 43.98 & 40.94  \\
        ViT-B~\cite{dosovitskiy2021vit} & 85.8 & 0.610 & 0.478 & 47.14 & 44.06 \\
        EfficientNetV2-S~\cite{tan2021efficientnetv2} (Ours) & 20.2 & \textbf{0.661} & \textbf{0.528} &  \textbf{43.01} & \textbf{39.96} \\
        \bottomrule
    \end{tabular}
    }
\end{table}

\noindent{\textbf{Vision Encoder.}}
\cref{table:vision_encoder} shows the ablation study on the vision encoder.
We compare EfficientNetV2-S~\cite{tan2021efficientnetv2} with ResNet50~\cite{he2016resnet}, which has a comparable number of parameters and is widely used in vision tasks, and with ViT-B~\cite{dosovitskiy2021vit}, which has recently been adopted in many vision tasks.
The results confirm that EfficientNetV2-S, which is commonly employed in hand pose estimation studies~\cite{ohkawa2023AssemblyHands, mucha2024inmyhands, mucha2026unseendomains}, is also well suited for our task.
As vision encoders continue to advance, our framework is expected to benefit from adopting newer encoders in the future.

\begin{table}[t]
    \centering
    \caption{\textbf{Comparison of Hand Detectors on N-HOT3D.} The best values are shown in \textbf{bold}, and the second-best values are \underline{underlined}.}
    \label{table:hand_detector}
    \setlength{\tabcolsep}{3pt}
    \scalebox{0.9}{
    \begin{tabular}{lccc}
        \toprule
        Detector & mAP@50 ($\uparrow$)  & mAP@50--95 ($\uparrow$) & FPS \\
        \midrule
        YOLO11~\cite{yolo11_ultralytics} & 0.699 & 0.375 & \textbf{221.03} \\
        RF-DETR~\cite{rf-detr} & \textbf{0.769}  & \underline{0.398} &  68.49 \\
        YOLO26~\cite{yolo26_ultralytics} (Ours) & \underline{0.750} & \textbf{0.407} & \underline{209.54}   \\
        \bottomrule
    \end{tabular}
    }
\end{table}

\noindent{\textbf{Hand Detector.}}
\cref{table:hand_detector} shows the ablation study on the hand detector.
We compare the adopted YOLO26 with its previous official version, YOLO11, and with RF-DETR, a Transformer-based detector widely used alongside YOLO.
Although RF-DETR achieves a higher mAP@50, YOLO26 is the most balanced model in terms of both accuracy and speed, as indicated by mAP@50--95 and FPS.

\begin{figure}[t]
  \centering
  \includegraphics[width=\columnwidth]{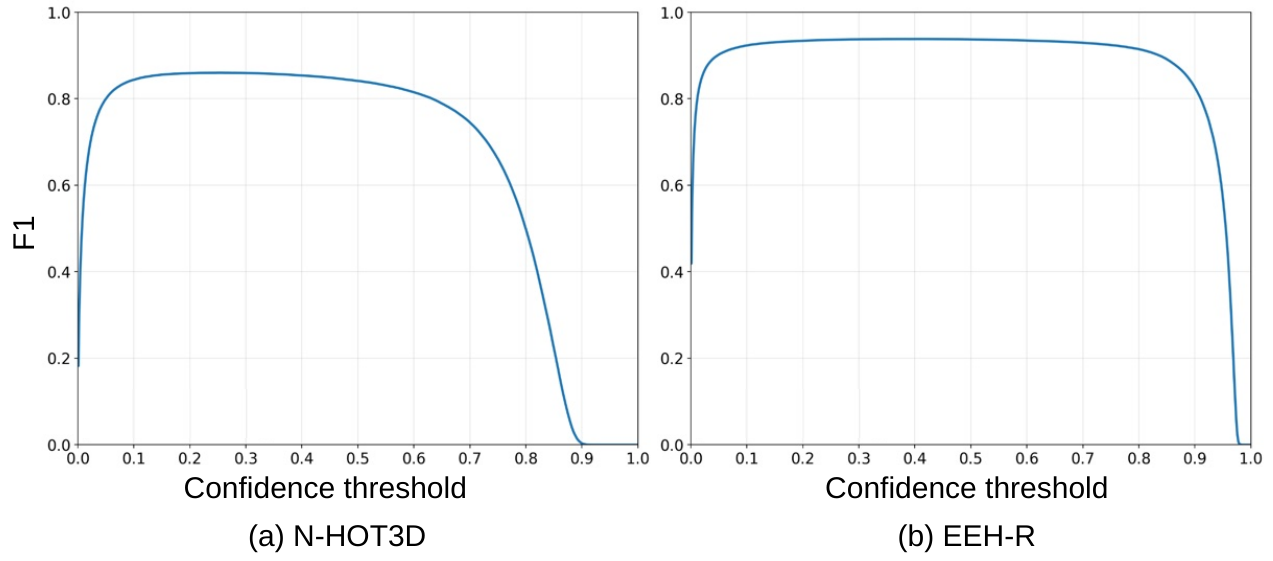}
  \vspace{-4mm}
  \caption{{F1 score of the Hand Detector at IoU~=~0.5 for varying confidence thresholds. (a)~N-HOT3D (synthetic). (b)~EEH-R (real).}}
  \label{fig:conf_threshold}
\end{figure}

\noindent{\textbf{Detection Threshold Analysis.}}
\cref{fig:conf_threshold} shows the F1 score at IoU~=~0.5 for varying confidence thresholds on the synthetic and real datasets.
On both datasets, the F1 score remains stable over a range of thresholds before dropping sharply beyond a certain point.
Since false positives introduce non-existent hands into the subsequent reconstruction stage, we empirically select a relatively high threshold that does not substantially degrade the F1 score, namely 0.5 for the synthetic dataset and 0.8 for the real dataset.

\noindent{\textbf{Adaptive Attention.}}
\cref{table:ablation_attention} shows the ablation study on the attention components in Adaptive Attention.
Self-attention mainly improves wrist-relative pose accuracy, while cross-attention improves RR-AUC, which evaluates the relative position between the two hands, by exchanging information across hands.
Combining both achieves the best performance on all metrics.

\begin{table}[t]
    \centering
    \caption{\textbf{Ablation study of attention components on N-HOT3D.} The best values are shown in \textbf{bold}.}
    \label{table:ablation_attention}
     \setlength{\tabcolsep}{3pt}
     \scalebox{0.8}{
    \begin{tabular}{cc|cccc}
        \toprule
        SelfAttn & CrossAttn & R-AUC ($\uparrow$) & RR-AUC ($\uparrow$) & MPJPE [mm] ($\downarrow$) & MPVPE [mm] ($\downarrow$) \\
        \midrule
        &  & 0.652 & 0.469 & 44.01 & 41.04\\
          \checkmark & & 0.657 & 0.517 & 43.37 & 40.34 \\
         & \checkmark  & 0.649 & 0.518 & 43.93 & 40.90 \\
          
        \checkmark & \checkmark & \textbf{0.661} & \textbf{0.528} & \textbf{43.01} & \textbf{39.96} \\
        \bottomrule
    \end{tabular}}
\end{table}

\noindent{\textbf{Mask Dilation Kernel Size.}}
\cref{table:dilation_kernel} shows the ablation study on the kernel
size used to dilate the masks estimated by the Hand Detector.
Hand mesh reconstruction performance is highest with a kernel size of $5\times5$ or $7\times7$.
Dilating the mask compensates for incomplete or partially missing segmentation results, whereas an excessively large kernel includes more non-hand regions, indicating that there is a limit to the benefit of dilation.
We adopt a kernel size of $7\times7$, which achieves the highest RR-AUC, as we prioritize the correct relative positioning of the two hands.

\begin{table}[t]
    \centering
    \caption{\textbf{Comparison of dilation kernel sizes on N-HOT3D.} The best values are shown in \textbf{bold}, and the second-best values are \underline{underlined}.}
    \label{table:dilation_kernel}
    \setlength{\tabcolsep}{3pt}
    \scalebox{0.85}{
    \begin{tabular}{lcccc}
        \toprule
        Method & R-AUC ($\uparrow$) & RR-AUC ($\uparrow$) & MPJPE [mm] ($\downarrow$) & MPVPE [mm] ($\downarrow$) \\
        \midrule
        w/o dilation & 0.636 & 0.467 & 44.84 & 41.71 \\
         kernel size 3 & 0.659 & 0.504 & 43.16 & 40.12 \\
         kernel size 5 & \textbf{0.667} & \underline{0.525} & \textbf{42.60} & \textbf{39.58} \\
         kernel size 7 & \underline{0.661} & \textbf{0.528} & \underline{43.01} & \underline{39.96} \\
         kernel size 9 & 0.644 & 0.520 & 44.30 & 41.16 \\
        \bottomrule
    \end{tabular}
    }
    % \vspace{-2mm}
\end{table}

\noindent{\textbf{Cross-Dataset Evaluation.}}
\cref{tab:cross_dataset} shows the quantitative results of the cross-dataset evaluation.
When the model is trained on N-HOT3D and evaluated on EEH-R, the performance drops substantially due to the domain gap between the two datasets.
However, when the model pre-trained on N-HOT3D is fine-tuned on EEH-R, it achieves slightly higher accuracy than the model trained on EEH-R alone.
This result indicates the potential of the synthetic dataset for pre-training.

\begin{table*}[t]
\centering
\caption{Cross-dataset evaluation. The best values are shown in \textbf{bold}.}
\label{tab:cross_dataset}
\begin{tabular}{ccc|cccc}
\hline
Train & Finetune & Test
& R-AUC ($\uparrow$) & RR-AUC ($\uparrow$) & MPJPE [mm] ($\downarrow$) & MPVPE [mm] ($\downarrow$)  \\
\hline
EEH-R & -- & EEH-R
& 0.691 & 0.551 & 34.184 & \textbf{32.163} \\
N-HOT3D & -- & EEH-R
& 0.079  & 0.038 & 137.229 & 126.446 \\
N-HOT3D & EEH-R & EEH-R
& \textbf{0.695} & \textbf{0.557} & \textbf{34.172} & 32.207 \\
\hline
\end{tabular}
\end{table*}

\noindent{\textbf{Impact of Dark-Scene Annotations on the Hand
Detector.}}
Since the number of manually annotated dark-scene masks is much smaller than that of the automatically generated well-lit masks, we analyze how the amount of dark-scene supervision affects the detector performance on EEH-R.
\cref{tab:eehr_cross_illumination} compares models trained on each illumination condition separately.
The model trained only on dark scenes performs worse under the dark condition than the model trained on both conditions, and its performance under the well-lit condition degrades substantially.
This indicates that combining the automatically generated well-lit annotations with the manual dark-scene annotations is an effective supervision strategy.

\begin{figure}[t]
  \centering
  \includegraphics[width=\columnwidth]{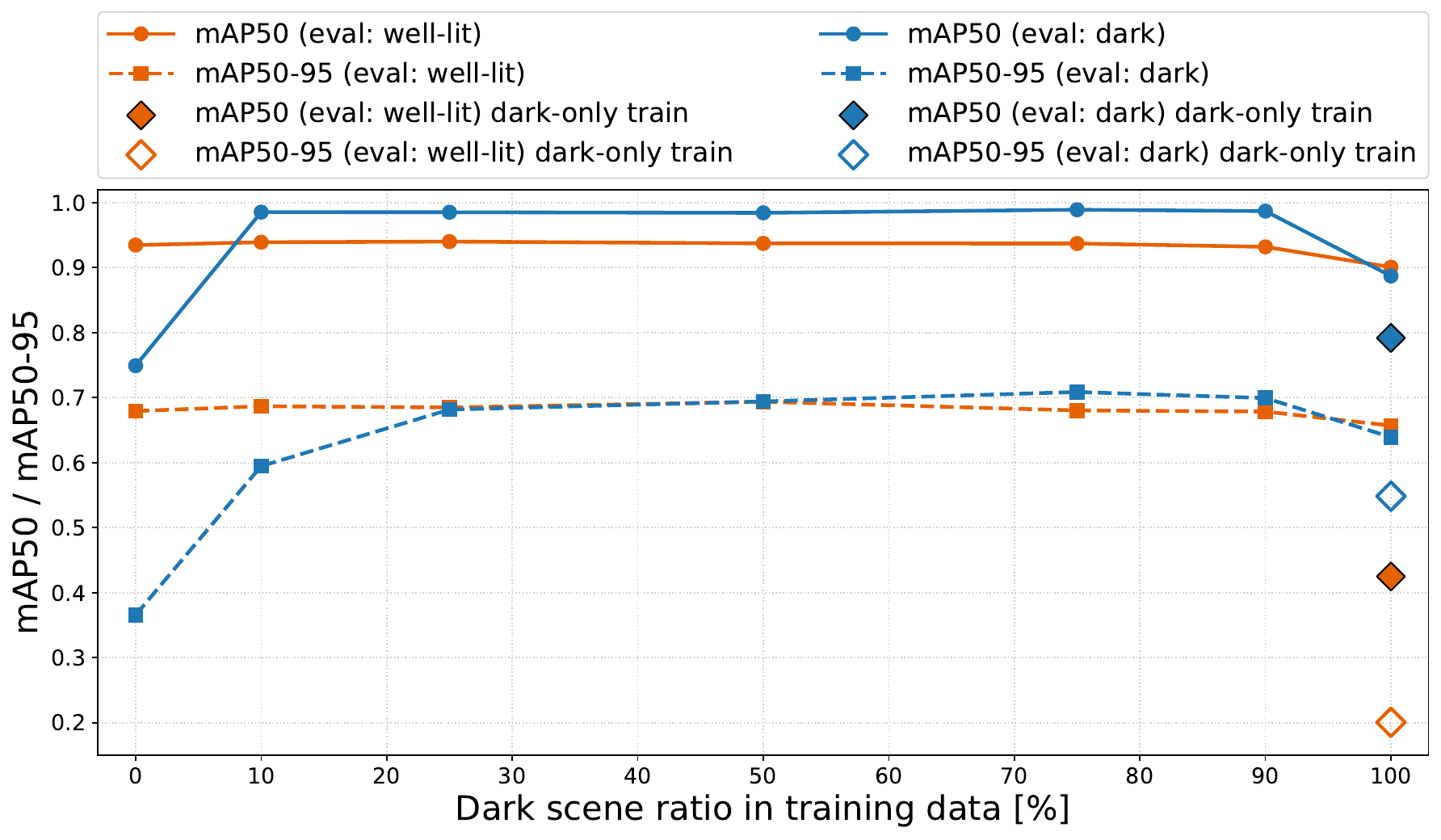}
  \vspace{-4mm}
  \caption{{Effect of the dark-scene data ratio on the performance of the Hand Detector on EEH-R. All well-lit annotations are used, while the ratio of dark-scene annotations in the training set is varied from 0\% to 100\%.}}
  \label{fig:dark_data_ratio}
\end{figure}

We further analyze how many dark-scene annotations are required.
We train the Hand Detector with all well-lit annotations in the training set while varying the ratio of dark-scene annotations from 0\% to 100\%, and evaluate the models separately under the well-lit and dark conditions.
As shown in \cref{fig:dark_data_ratio}, the performance under the well-lit condition remains almost constant for both metrics regardless of the dark-scene data ratio.
Under the dark condition, mAP@50--95 increases with the ratio and saturates once the ratio reaches approximately 25\% of the dark training set.
This demonstrates that adding only a small number of manually annotated dark frames to the well-lit data is sufficient for reliable detection in dark scenes.

\begin{table}[t]
    \centering
    \caption{\textbf{Cross-illumination evaluation of the Hand Detector on EEH-R.} Each model is trained and tested under different illumination conditions. The best values for each test condition are shown in \textbf{bold}.}
    \label{tab:eehr_cross_illumination}
    \setlength{\tabcolsep}{6pt}
    \scalebox{0.9}{
        \begin{tabular}{l l cc}
        \toprule
        Train & Test & mAP@50 ($\uparrow$) & mAP@50--95 ($\uparrow$) \\
        \midrule
        
        Well-lit & Well-lit & \textbf{0.935} & \textbf{0.679} \\
        Dark & Well-lit & 0.425 & 0.201 \\
        All & Well-lit & 0.901 & 0.657 \\
        \hline
        Well-lit & Dark & 0.749 & 0.366 \\
        Dark & Dark & 0.792 & 0.548 \\
        All & Dark & \textbf{0.887} & \textbf{0.639} \\
        \hline
        Well-lit & All & \textbf{0.934} & \textbf{0.678}\\
        Dark & All & 0.427 & 0.202 \\
        All & All & 0.928 & 0.655 \\
        \bottomrule
        \end{tabular}
    }
    % \vspace{-2mm}
\end{table}

\subsection{Failure Analysis}
\label{ssec:failure}
\begin{figure}[t]
  \centering
  \includegraphics[width=\columnwidth]{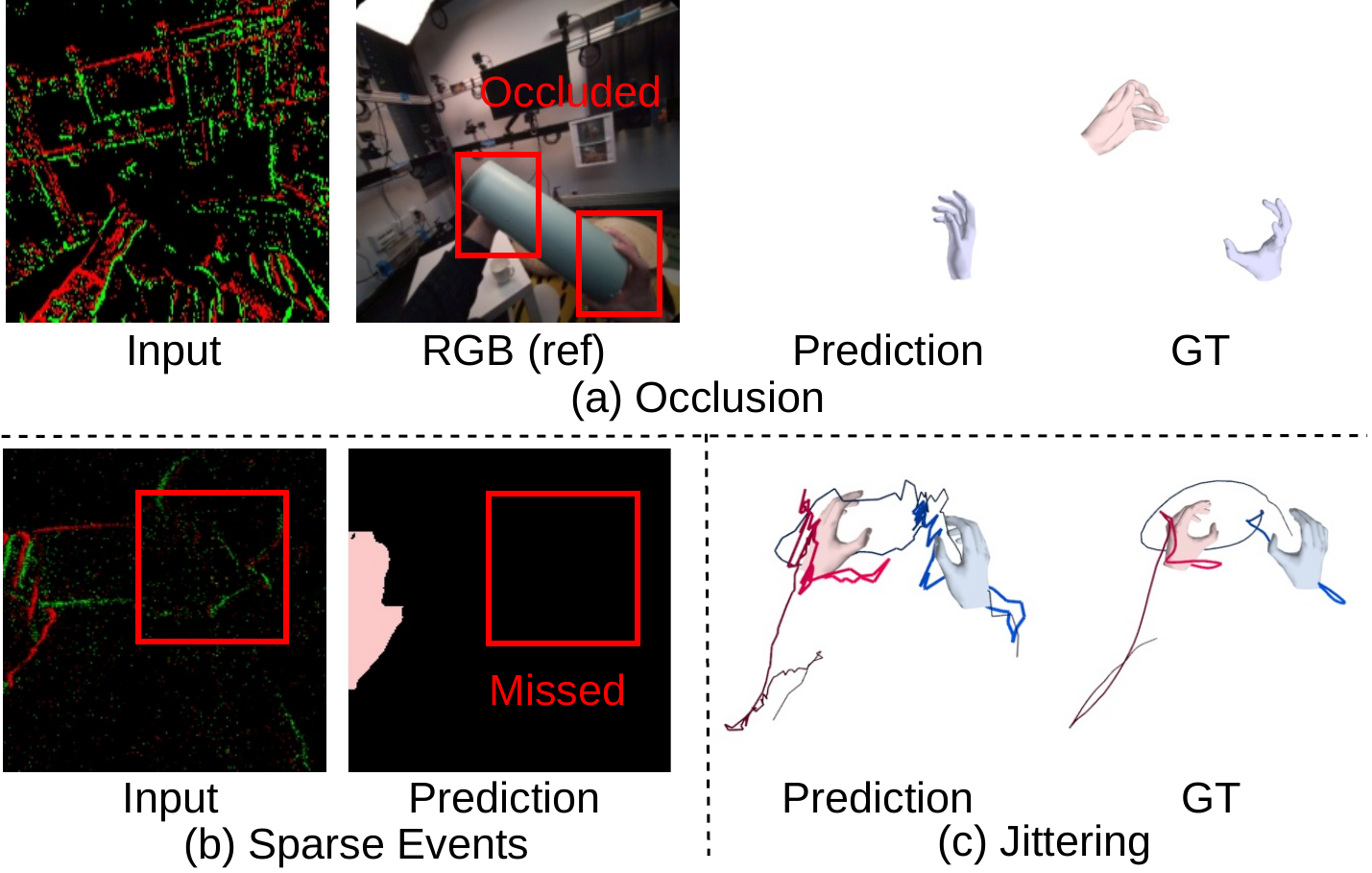}
  \vspace{-4mm}
  \caption{Failure cases of the proposed method.
  (a)~Missed detection and inaccurate pose estimation under severe occlusion by a held object.
  (b)~Missed detection due to sparse events when the hand remains nearly static.
  (c)~Temporal jittering in the predicted trajectory due to
frame-by-frame estimation without temporal modeling.
Red and blue lines indicate the wrist trajectories of the left
and right hands, respectively.
}
  \label{fig:failure}
\end{figure}

We identify three major failure patterns of the proposed method, as illustrated in \cref{fig:failure}.

\noindent\textbf{Occlusion.}
When a hand is occluded by an object, the Hand Detector may fail to detect it. 
Even when the hand is detected, accurately reconstructing the occluded parts, such as the fingertips, remains difficult when they are hidden by the object, as shown in \cref{fig:failure}~(a).
Since hands frequently interact with surrounding objects in daily activities, explicitly modeling object shape and hand--object contact is a promising direction.
While current event-based approaches primarily focus on hand regions, modeling hand--object interactions remains largely unexplored.
Inspired by RGB-based methods~\cite{park2022handoccnet, lin2023harmonious}, incorporating distinct feature representations for both hand and object regions is a promising direction for improving reconstruction accuracy in complex interaction scenarios.

\noindent\textbf{Sparse Events.}
Since an event camera only responds to brightness changes, events become sparse when the hand and head remain nearly static.
In such cases, the hand may not be detected, as shown in \cref{fig:failure}~(b).
This effect is more pronounced in low-light environments, where the detection performance is lower than in well-lit conditions, as indicated by the mAP in \cref{tab:mask_results}.
Accumulating past events or introducing temporal processing would help mitigate this failure.

\noindent\textbf{Jittering.}
Since the proposed method operates on a frame-by-frame basis, the reconstructed hand meshes may exhibit temporal jittering across consecutive frames.
As shown in the wrist trajectories in \cref{fig:failure}~(c), the predicted motion is less smooth than the ground truth. 
Note that such jittering is not specific to event-based methods, as frame-based RGB methods such as HaMeR~\cite{pavlakos2024hamer} and WiLoR~\cite{potamias2025wilor} also suffer from it.
Leveraging the temporal nature of event streams and extending the framework to 4D hand mesh reconstruction is an important direction for future work.
Recent approaches such as HaPTIC~\cite{ye2026prediciting4d}, which predicts coherent 4D hand trajectories from monocular videos by fusing temporal information across frames with attention and by directly estimating inter-frame depth changes rather than per-frame absolute depth, suggest promising strategies to resolve this issue.

%================================================================

%================================================================
\section{Conclusion}
\label{sec:conclusion}
In this paper, we propose EventEgoHands++, a framework for event-based egocentric 3D hand mesh reconstruction.
To address the challenges of background noise and the inability of prior work to distinguish between hands, we introduce a Hand Detector that performs instance-level hand detection by jointly estimating bounding boxes and masks.
Furthermore, to overcome the performance degradation caused by fixed attention that ignores hand visibility,
we also propose Adaptive Attention that dynamically adjusts the attention process depending on the presence of detected hands.
In addition, we extend the synthetic N-HOT3D dataset with
refined masks and newly added bounding-box annotations, and newly
construct the real-world EEH-R dataset.
Experiments on both synthetic and real datasets demonstrate that the proposed method significantly improves reconstruction performance compared with existing approaches.
We hope that this work contributes to advancing event-based hand reconstruction research and facilitates future studies on egocentric event-based vision systems.

\newpage

\bibliographystyle{unsrt}
\bibliography{access_refs}

\EOD

\end{document}